\pdfoutput=1

\documentclass[11pt]{article}

\usepackage{acl}

\usepackage{times}
\usepackage{latexsym}
\usepackage{amsmath}
\usepackage{xcolor}      
\usepackage{ragged2e}
\usepackage{makecell}
\usepackage{tcolorbox}
\usepackage{CJKutf8}
\usepackage[utf8]{inputenc}
\usepackage{enumitem}

\usepackage{cleveref}
\usepackage[T1]{fontenc}

\usepackage{booktabs,tabularx,colortbl,multirow,array,makecell}

\usepackage{microtype}

\usepackage{inconsolata}
\usepackage{algorithm}  
\usepackage{algpseudocode}

\usepackage{fancyhdr}
\usepackage{graphicx}
\usepackage{float}
\usepackage{amsfonts}
\usepackage{acronym}
\usepackage{balance}
\usepackage{xspace}
\usepackage{setspace}
\usepackage{subcaption}
\usepackage{arydshln}
\usepackage{amssymb,graphicx} 

\title{\textsc{CLaw}: Benchmarking Chinese Legal Knowledge in Large Language Models -- A Fine-grained Corpus and Reasoning Analysis}

\author{
  Xinzhe Xu\textsuperscript{1,2}\thanks{Work was done during an internship at LLM-Core Xiaomi. Email: \texttt{ypxxz2019@pku.edu.cn}.} \quad
  Liang Zhao\textsuperscript{2} \quad
  Hongshen Xu\textsuperscript{2} \quad
  Chen Chen\textsuperscript{2} \\
  \textsuperscript{1}Peking University \quad
  \textsuperscript{2}LLM-Core Xiaomi
}

\begin{document}
\begin{CJK}{UTF8}{gbsn} 
\maketitle

\begin{abstract}
Large Language Models (LLMs) are increasingly tasked with analyzing legal texts and citing relevant statutes, yet their reliability is often compromised by general pre-training that ingests legal texts without specialized focus, obscuring the true depth of their legal knowledge. This paper introduces \textsc{CLaw}, a novel benchmark specifically engineered to meticulously evaluate LLMs on Chinese legal knowledge and its application in reasoning. \textsc{CLaw} comprises two key components: (1) a comprehensive, fine-grained corpus of all 306 Chinese national statutes, segmented to the subparagraph level and incorporating precise historical revision timesteps for rigorous recall evaluation (64,849 entries), and (2) a challenging set of 254 case-based reasoning instances derived from China Supreme Court curated materials to assess the practical application of legal knowledge. Our empirical evaluation reveals that most contemporary LLMs significantly struggle to faithfully reproduce legal provisions. As accurate retrieval and citation of legal provisions form the basis of legal reasoning, this deficiency critically undermines the reliability of their responses. We contend that achieving trustworthy legal reasoning in LLMs requires a robust synergy of accurate knowledge retrieval—potentially enhanced through supervised fine-tuning (SFT) or retrieval-augmented generation (RAG)—and strong general reasoning capabilities. This work provides an essential benchmark and critical insights for advancing domain-specific LLM reasoning, particularly within the complex legal sphere.

\end{abstract}

\section{Introduction}

\begin{figure}[t]
\centering
\includegraphics[width=1.0\linewidth]{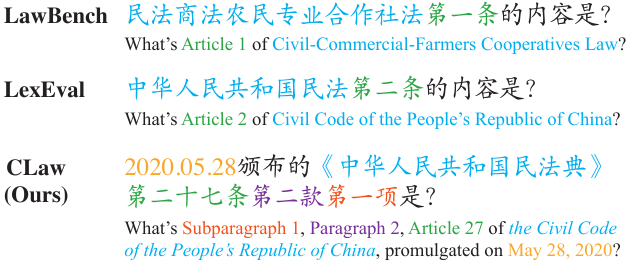}
\includegraphics[width=1.0\linewidth]{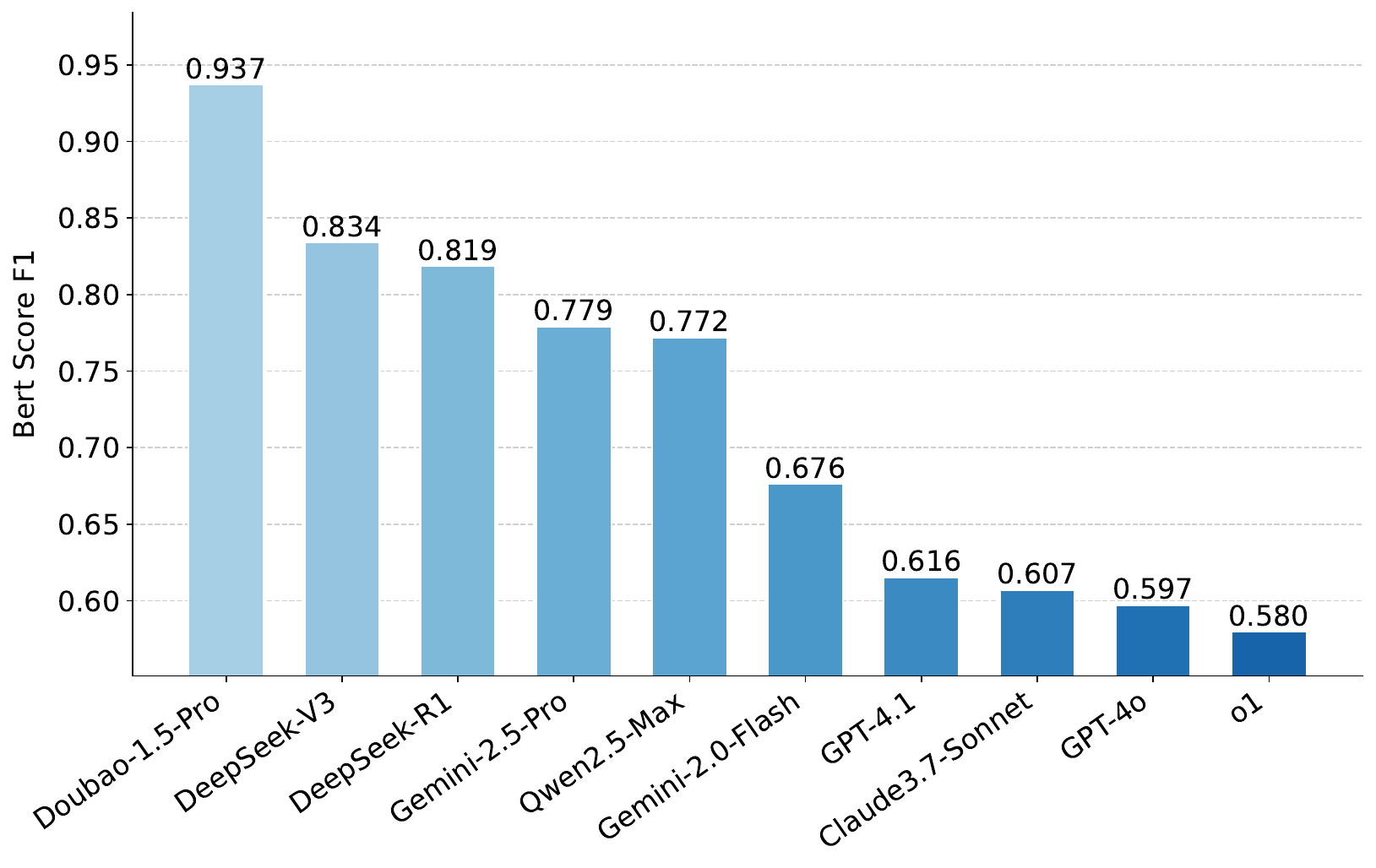}
\vspace{-2em}
\caption{(top) Comparison between our query set (\textsc{CLaw}) and the queries in LawBench~\citep{fei2023lawbench} and LexEval~\citep{li2024lexeval}, which often cite statutes imprecisely and omits temporal information, whereas our queries follow the subparagraph‑level format used in judicial opinions. (bottom) BERTScore of ten LLMs when asked to recite \emph{Civil Code of the People's Republic of China}, one of the most frequently used laws.}
\label{fig:fig1}
\vspace{-1em}
\end{figure}

Large Language Models (LLMs) have demonstrated remarkable capabilities, approaching expert-level performance in intricate domains such as coding~\citep{wang2024openhands} and mathematics~\citep{glazer2024frontiermath}. This success has catalyzed significant interest in deploying LLMs in high-stakes fields like medicine~\citep{moor2023foundation}, law~\citep{zhou2024lawgpt}, and economics~\citep{axtell2025agent}. However, effective deployment in such specialized areas demands more than superficial understanding; it necessitates genuine mastery of domain-specific knowledge and the capacity for sound reasoning based on that knowledge. A core assertion of this paper is that effective domain-specific reasoning is not merely an extension of general reasoning capabilities. Instead, it emerges from a crucial synergy: comprehensive, accurate domain knowledge coupled with robust general thinking abilities. This distinction is pivotal because current LLMs, typically pre-trained on broad, general corpora, may encounter and process domain-specific texts (like legal statutes) without achieving the necessary depth of knowledge mastery. Consequently, the reliability of their subsequent reasoning remains uncertain, irrespective of their inherent general cognitive skills.

This paper investigates two fundamental questions through this lens: (1) Do state-of-the-art LLMs possess reliable, fine-grained knowledge of law, specifically for the precise statute recall essential in judicial practice (the knowledge mastery component)? (2) How effectively can these LLMs apply this legal knowledge by leveraging their general thinking abilities in practical case analysis (the reasoning application component)?

We focus our investigation on Chinese law, a domain that presents both a critical societal need for AI assistance and an ideal testbed for evaluating these two distinct yet intertwined components. The sheer volume of litigation in China, with millions of judgments published annually~\citep{hu2024does}, highlights the urgent demand for reliable AI-driven legal tools. Its civil-law system, where judicial decisions primarily hinge on written statutes rather than case precedents, offers a well-defined and structured body of knowledge—a prerequisite for rigorously testing knowledge mastery. Furthermore, the moderate pace of amendments to Chinese statutes necessitates that LLMs accurately distinguish between multiple historical versions of the same legal article, providing a stringent test of knowledge precision and currency. Such precision is foundational; without robust knowledge mastery, any downstream reasoning, however sophisticated its general approach, will inherently be flawed.

Despite the critical importance of accurate knowledge recall and its subsequent application, existing benchmarks often categorize recall as a ``basic'' task~\citep{fei2023lawbench} and utilize queries with vague or temporally ambiguous references (Fig~\ref{fig:fig1}(top)). This overlooks the role of precise knowledge mastery as an independent pillar of domain expertise. Moreover, evaluating the practical application of this knowledge—the efficacy of general reasoning operating upon this specialized knowledge base—remains a significant challenge.

To address these limitations, we introduce \textsc{CLaw}, a dual-faceted benchmark meticulously designed to dissect and evaluate these two components. Firstly, to assess knowledge mastery, \textsc{CLaw} features a comprehensive corpus encompassing all 306 Chinese national statutes, including every available historical version (424 distinct statute versions in total). This corpus is segmented to the subparagraph level and incorporates precise revision timesteps, enabling fine-grained queries that mirror the citation format used in actual judicial opinions. This results in 64,849 distinct entries for statutory knowledge evaluation. Secondly, to assess the application of general thinking abilities to this domain knowledge, \textsc{CLaw} introduces a case-based reasoning task. This component, built upon 254 representative and challenging cases curated by the China Supreme Court (complete with judicial reasoning for reference), evaluates how well LLMs can analyze case descriptions, identify relevant legal issues, and apply statutory knowledge through logical reasoning to derive conclusions. We also propose an LLM-as-a-judge pipeline for the faithful and automated evaluation of this reasoning task.

Our empirical results from the statutory recall benchmark, targeting the knowledge mastery aspect, are concerning. Even leading models like the GPT series (o1, GPT-4o, GPT-4.1) frequently cite repealed or hallucinated provisions when queried about Chinese law (Figure~\ref{fig:fig1}(bottom)). This fundamental deficiency in knowledge inevitably compromises any subsequent attempts at higher-level legal reasoning, regardless of an LLM's general cognitive prowess. Preliminary analysis of our case-based reasoning task further suggests that both deep domain knowledge and strong general thinking abilities are indispensable for tackling real-world legal tasks. This study underscores that before LLMs can be reliably deployed in high-stakes fields like law, their domain-specific knowledge mastery must be rigorously assessed and substantially augmented. Only then can their general reasoning abilities be effectively leveraged to produce dependable domain-specific applications. Our contributions are:
\begin{itemize}
\item Introduced \textsc{CLaw}, a pioneering benchmark for Chinese legal domain, uniquely featuring a subparagraph-level, historically versioned corpus of all national statutes (64,849 entries) and challenging case-based reasoning tasks.
\item Revealed critical deficiencies in current leading LLMs regarding precise legal knowledge recall, demonstrating that even advanced models struggle with factual accuracy, thereby undermining their legal reasoning reliability.
\item Provided empirical evidence supporting the thesis that trustworthy domain-specific reasoning in LLMs necessitates a synergistic combination of deep, accurate knowledge mastery and robust general reasoning capabilities.
\end{itemize}

\section{Related Work}
\label{sec:related_work}

\paragraph{Knowledge memorization of LLMs}
LLMs inherently memorize training data, a double-edged sword offering factual recall but risking leakage of sensitive information. This phenomenon, influenced by model scale and data repetition~\citep{menta2025analyzing}, can be detected via methods like extraction attacks~\citep{aditya2024evaluating} and MIAs~\citep{meeus2024sok}. Balancing utility and safety remains a challenge~\citep{li2025memorization}.

\paragraph{Legal benchmarks for LLMs}
The legal domain's specialized language and complex reasoning necessitate domain-specific benchmarks. Key benchmarks include LegalBench (English \cite{guha2023legalbench}), the multilingual LEXTREME~\cite{niklaus2023lextreme}, and jurisdiction-specific ones like oab-bench (Brazilian Portuguese~\citep{pires2025automatic}). These often cover tasks like legal text classification, summarization, question answering, and to some extent, reasoning. 

\paragraph{Chinese Legal LLMs}
China's distinct legal system has spurred development of specialized LLMs and benchmarks. Notable benchmarks are LawBench \cite{fei2023lawbench} and LexEval~\citep{li2024lexeval}, which uniquely includes ethical evaluations. Leading models like Lawyer LLaMA \cite{huang2023lawyer}, ChatLaw~\citep{cui2023chatlaw}, and InternLM-Law~\citep{fei2024internlm} are trained on diverse Chinese legal texts and filtered consultation data, aiming to effectively navigate the local legal landscape. \textsc{CLaw} builds upon this by providing an even more granular dataset for statutory knowledge and a dedicated component for case-based reasoning.

\paragraph{AI-assisted Legal Case Analysis}
Early efforts in AI and law centered on emulating legal reasoning through expert systems, particularly for interpreting written statutes where rules are explicit \cite{Rissland1990}. This evolved into the development of more sophisticated computational models aimed at representing legal knowledge and formalizing the argumentation processes involved in applying statutes to case facts \cite{PrakkenSartor2015}. Such models often treat legal texts, including statutes and contracts, as programs, allowing for systematic analysis akin to software engineering methods to enhance AI-driven statutory interpretation \cite{Padhye2024}. While LLMs have shown promise in processing and analyzing legal documents \cite{Aletras2016}, their application to statutory reasoning requires robust mechanisms for logical consistency, explainability, and handling the nuances of legal language to ensure trustworthy outcomes \cite{Alshamsi2025}. The structured nature of statutory law provides a fertile ground for AI systems that can perform deliberate, logical reasoning, yet ensuring transparency and the ability to scrutinize AI-driven conclusions remains a critical research focus.

\section{\textsc{CLaw} Benchmark: Corpus and Tasks}

\subsection{Foundational Statutory Knowledge: The \textsc{CLaw} Corpus}

To begin with, we would like to first discuss the motivation of curating a legal knowledge corpus through a debate, since alternative view may think that many LLMs provide the API to ``search online'' for these knowledge.

\subsubsection{A Preliminary Debate: SFT or RAG for Legal Knowledge?}

In judicial adjudication, Chinese law typically requires judges to reason from the basic facts, identify the specific statutory provisions, and then reach a ruling. Unlike other knowledge, statutes and regulations change relatively slowly and carry greater authority. Consequently, whether to embed statutory provisions directly into a large model’s knowledge base has long been a matter of debate.
Another intuitive approach is to use retrieval‑augmented generation (RAG), letting the model search online and pull the relevant articles based on the case facts~\citep{wei2025browsecomp}. Yet our preliminary experiments show that current LLMs are still weak at retrieving legal provisions in~\Cref{app.pre_expr}. The main reasons are: (1) today’s search methods rely on shallow similarity matching, whereas statute retrieval demands precise understanding and reasoning; (2) there has been no carefully curated corpus of statutes—multiple versions circulate online, and search results easily mix up timeframes.
Our work releases the open‑source \textsc{CLaw} statutory corpus to mitigate the second problem, but the first—insufficient retrieval capability—remains a challenge. Although this study examines how well LLMs memorize statutes, it is not a call to embed all legal knowledge into the model. Instead, we seek to show the research community how much domain knowledge LLMs currently command, so they can judge whether the LLMs can be trusted for legal reasoning. We also point out that to enhance in‑domain reasoning, one must rely on supervised fine‑tuning (SFT) or RAG to ensure the correctness of legal knowledge, and we hope this spurs further work on solving the distinct issues each approach faces. The \textsc{CLaw} corpus provides a high-quality resource for both SFT approaches and for building more robust RAG systems.

\subsubsection{Fine-grained Corpus Construction}

We initially crawled a comprehensive set of 453 statutes from the \emph{National Database of Laws and Regulations}\footnote{\url{https://flk.npc.gov.cn/}}.
Adhering strictly to the legal definition of \emph{law}, we removed non‑statutory documents, such as State Council regulations and other instruments issued without a Presidential Order.
Each statute is stored and identified by its title and version date, capturing its full revision history. This temporal precision is crucial for legal accuracy.
For every article, we further parsed its structure to identify whether it contains multiple paragraphs(款) and subparagraphs(项), as judicial opinions often require citation to the exact subparagraph if applicable. This fine-grained, \emph{subparagraph-level segmentation} combined with \emph{precise revision timesteps} constitutes a key feature of our corpus. This initial parsing yielded 64,849 entries. These entries are saved using a hierarchical structure: ⟨law-name⟩−⟨version⟩−⟨article⟩−⟨paragraph⟩−⟨subparagraph⟩,
where the paragraph or subparagraph field is left empty if not applicable. This detailed structure is essential for precise evaluation and for developing LLMs that can understand and reference law at the required level of detail.

\subsubsection{Statutory Knowledge Retrieval Tasks}
To evaluate law recitation performance based on the \textsc{CLaw} corpus, we designed two tasks:
\begin{itemize}
    \item \textbf{ID Retrieval}: Given the law's version, the model must retrieve the corresponding article, paragraph, and subparagraph index.
    \item \textbf{Content Retrieval}: Given the law's article, paragraph, and subparagraph index, the model must recite the corresponding content.
\end{itemize}
These tasks directly test the model's ability to accurately recall and locate specific legal provisions from the fine-grained corpus.

\subsection{Case-Based Legal Reasoning Task}
\label{sec.case-based-task}
To evaluate the legal reasoning capabilities of LLMs in practical scenarios, we curated a specialized test set from representative Guiding Case (指导性案例) issued by the Supreme People's Court (SPC) of China.\footnote{\url{https://www.court.gov.cn/shenpan/gengduo/77.html}} Legal reasoning processes are inherently distinct from those in disciplines like coding or math, where singular, verifiable ground truth outcomes are often attainable. Evaluating the relative merits of different legal arguments presents considerable challenges because judicial decisions can be subject to varied interpretations, may possess flaws, and are frequently appealed and potentially overturned by higher courts.

To navigate these complexities, our test set exclusively employs Guiding Case. These cases are meticulously selected and published by the SPC to promote the uniform application of law and to provide authoritative references for judicial practice nationwide. Consequently, the analyses and reasoning presented in these documents exemplify a high standard of legal interpretation within the Chinese legal system. Each selected Guiding Case instance offers a structured and comprehensive summary, encompassing:

\begin{itemize}
    \item \textbf{Case Details:} A concise overview of the factual background of the case.
    \item \textbf{Adjudication Outcomes:} The final decision or judgment rendered by the court.
    \item \textbf{Focus of Dispute:} The central legal and/or factual questions deliberated by the court.
    \item \textbf{Judicial Reasoning:} The detailed rationale underpinning the court's decision, including its interpretation of legal principles and their application to the case's facts.
    \item \textbf{Relevant Legal Articles:} Specific legal provisions (cited to the article, paragraph, and subparagraph level, where applicable) that form the basis of the judgment.
\end{itemize}

This structured format facilitates a targeted evaluation of an LLM's capacity for legal reasoning. Specifically, we assess an LLM's ability to identify and analyze the primary ``focus of dispute'' – which can often be distilled into a core question – and compare its generated response against the authoritative ``judicial reasoning'' provided in the Guiding Case. A complete illustrative example can be found in~\Cref{app:prompts}. While several Chinese legal benchmarks for LLMs have emerged (as discussed in~\Cref{sec:related_work}), to the best of our knowledge, this work is the first to leverage these high-quality, authoritative data sources for this purpose.

\paragraph{Dataset Curation and Evaluation Methodology.}
The SPC continuously updates Guiding Case, typically every 1-6 months. Our dataset includes all 256 cases published up to April 8, 2025. After removing two ineffective cases, our final dataset comprises 254 cases. To standardize the task, we employed GPT-4o to rephrase each case's descriptive ``focus of dispute'' into a direct question (e.g., transforming ``the dispute focus of this case is whether A or B'' into ``Based on the case details, is it A or B?''). For evaluation, we use a detailed rubric, provided in~\Cref{app:prompts}, for a judge LLM to rate the responses. Developed collaboratively by legal and LLM experts, the rubric incorporates five critical aspects of legal reasoning essential for actual judicial decision-making: \emph{reasoning rigor}, \emph{knowledge accuracy}, \emph{conciseness}, \emph{logical structure}, and \emph{clarity}. The judge LLM is required to give an overall rating (0-100) and give \emph{Good}, \emph{Normal}, \emph{Bad} judgment on the five dimensions.
This automated evaluation method demonstrates strong agreement with human assessments (Pearson correlation $r=0.82$, inter annotator agreement correlation $r_{human}=0.96$), which was established by having three legal experts annotate LLM responses for a subset of 20 cases.

\section{Benchmarking Knowledge Recall}
\subsection{Experimental Setup}

\begin{figure}[t]
    \centering
    \includegraphics[width=1.0\linewidth]{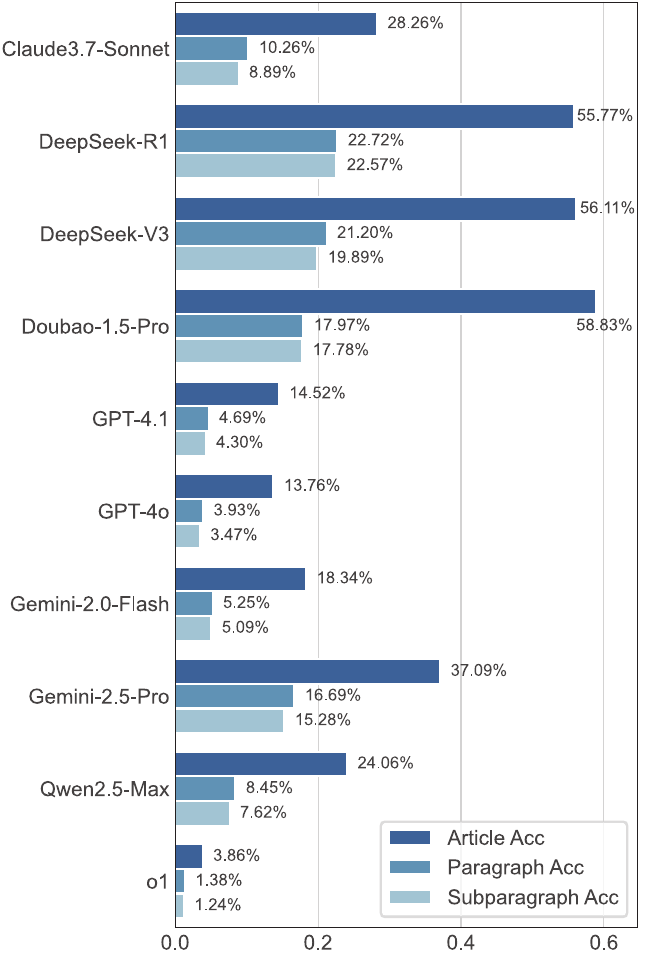}
    \caption{Three nested metrics on the ID Retrieval: Article Acc (dark blue, correct article index), Paragraph Acc (medium blue, correct article+paragraph indices), and Subparagraph Acc (light blue, correct article+paragraph+subparagraph indices). Results are averaged over the test set; higher values indicate better ability to recall the exact location of a cited statute.}
    \label{fig:acc}
\end{figure}

\paragraph{LLM set.}
We evaluate ten LLMs, including prominent models from China such as DeepSeek-R1~\citep{guo2025deepseekr1} and DeepSeek-V3~\citep{deepseekv3_2024}, Qwen2.5-Max~\citep{qwen2025max}, Doubao-1.5-pro~\citep{doubao2025pro}. We also benchmarked leading global models including o1~\citep{openai2024o1systemcard}, GPT-4o~\citep{openai2024gpt4o} and GPT-4.1~\citep{openai2025gpt41}, Gemini-2.5-Pro~\citep{google2025gemini25pro} and Gemini-2.0-Flash~\citep{google2025gemini2flash}, and Claude-3.7-Sonnet~\citep{anthropic2025claude37}. Note that our evaluation focuses on general-purpose LLMs and does not include models specifically fine-tuned for the legal domain.

\paragraph{Evaluation Metrics.}
For the ID Retrieval task, we employ hierarchical accuracy~\citep{silla2007review}. Article accuracy measures the percentage of correctly predicted article indices. Paragraph accuracy requires correct prediction of both article and paragraph indices. Subparagraph accuracy necessitates correct prediction of article, paragraph, and subparagraph indices. For the Content Retrieval task, we assess similarity between retrieved and ground truth content using ROUGE~\citep{lin-2004-rouge}, BLEU~\citep{papineni-etal-2002-bleu}, Edit Distance~\citep{przybocki-etal-2006-edit}, and BERT Score~\citep{bert-score}. Please refer to~\Cref{app:metrics} for a detailed introduction of the metrics.

\subsection{Results and Analysis of Statutory Knowledge Recall}
\label{sec:results_recall} 

\paragraph{Overall performance.} Due to space limitations, we postpone detailed Content Retrieval results to Appendix~\ref{app:metrics}. \Cref{fig:acc} reveals that ID Retrieval is far from a solved problem.
Even at \emph{article} level---the coarsest of the three granularities---the best-performing Doubao-1.5-Pro resolves the correct citation only \textbf{58.8\%} of the time.
Performance deteriorates sharply as finer detail is required; averaged across LLMs, \emph{paragraph}-level accuracy is roughly one-third of \emph{article}-level accuracy.

\begin{figure}[t]
    \centering
    \includegraphics[width=1.0\linewidth]{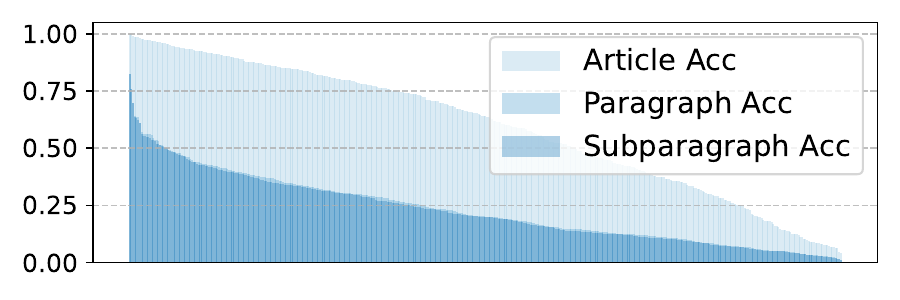}
    \caption{Per‑statute best‑LLM ID‑retrieval accuracy. Accuracy drops sharply beyond the article level, with paragraph and sub‑paragraph curves nearly overlapping.}
    \label{fig:distribution}
\end{figure}

\paragraph{Chinese oriented LLMs dominate statutory recall.}
The four Chinese-origin models secure the top four ranks in article-level accuracy:
Doubao-1.5-Pro (58.8\%), DeepSeek-V3 (56.1\%), DeepSeek-R1 (55.8\%), and Qwen-2.5-Max (24.1\%).
However, this ranking shifts for fine-grained retrieval: DeepSeek-R1 surpasses Doubao by \textbf{+4.75\%} on Paragraph Accuracy and \textbf{+4.79\%} on Sub-paragraph Accuracy, indicating a stronger understanding of exact statute content rather than mere memorization of the surface article structure.
The GPT series (GPT-4.1, GPT-4o, o1) trails significantly behind; \textbf{notably, the least effective o1 often responds with disclaimers of uncertainty instead of attempting an answer} (e.g., responding with ``I cannot tell you the answer''). We will incorporate an \emph{abstention rate} metric to quantify this in future work.
Gemini-2.5-Pro and Claude-3.7-Sonnet emerge as competitive non-Chinese LLMs, though they still lag behind their Chinese counterparts. This suggests that increased model size alone is insufficient for memorization, and factors like more intensive data repetition during training may play a crucial role.

\paragraph{Fine-grained difficulty in recall.} As illustrated in \Cref{fig:acc}, the task of ID Retrieval becomes significantly more challenging as the required granularity increases. Paragraph and subparagraph accuracies are substantially lower, with \Cref{fig:distribution} indicating that for many statutes, even the best-performing LLMs struggle to identify content beyond the article level (mostly below 50\% Acc). This demonstrates that while LLMs might sometimes recall the general vicinity of a legal provision (the article), they largely fail to grasp or retrieve the finer structural details (paragraph and subparagraph) essential for accurate legal citation. In Content Retrieval, a parallel observation is that the performance of reciting articles without (sub)paragraphs is significantly better than articles with (sub)paragraphs.

\paragraph{Error analysis in statutory recall.} A detailed analysis of common errors, exemplified by the best-performing model DeepSeek-R1 (see Appendix~\ref{app.r1} for a full breakdown), reveals several critical failure patterns:
(1) \textbf{Addition of irrelevant or fabricated content}, where models incorporate text from other clauses or articles, make minor alterations that change legal meaning, significantly rewrite provisions by mixing unrelated texts, or generate entirely non-existent legal content.
(2) \textbf{Incorrect versioning or indexing}, such as providing content from a different (often outdated) version of the law than requested, or citing the wrong article, paragraph, or subparagraph number for the provided text.
(3) \textbf{Outright refusal to answer}, where models inappropriately claim inability to access or provide the requested legal text due to their policy, even for publicly accessible statutes.
These errors highlight systemic issues in accurately recalling precise legal information.

\section{Benchmarking Legal Case Reasoning}

\subsection{Experimental Setup}

We evaluate the same set of 10 LLMs used as our candidate models. For evaluation, we employ {Gemini-2.5-Pro} and {DeepSeek-R1} as judge LLMs, selected for their strong general reasoning capabilities and superior understanding of legal knowledge, as demonstrated in our knowledge retrieval tasks. Given that both the rubric and reference answers are provided (introduced in~\Cref{sec.case-based-task}), the evaluation process is relatively objective. The two judge LLMs produce identical rankings and exhibit a high degree of agreement, with a Pearson correlation of $r = 0.98$ on absolute ratings. The results are presented in \Cref{tab:reasoning_performance,tab:reasoning_performance_deepseek_r1_judge}.
\begin{table*}[t]
\caption{Legal Case Reasoning performance as evaluated by an LLM judge (Gemini-2.5-Pro). Results judged by DeepSeek-R1 are available in~\Cref{app.llm_judge}, showing a Pearson Correlation of $r=0.98$ for overall rating. The maximum score for Overall Rating is 100, while five individual aspects are scored out of 20. Note that the overall rating is not a simple sum of these five aspects. For the aspects, qualitative ratings of \emph{Good}, \emph{Normal}, \emph{Bad} were converted to scores of 20.0, 10.0, and 0.0, respectively. All presented results are an average across 254 cases.
}
\label{tab:reasoning_performance} 
\centering 
\resizebox{\textwidth}{!}{%
\begin{tabular}{l>{\columncolor{cyan!20}}cccccc}
\toprule
        \textbf{Model Name} &  \textbf{Overall Rating} &  \textbf{Reasoning} &  \textbf{Knowledge} &  \textbf{Structure} &  \textbf{Clarity} &  \textbf{Conciseness} \\
        \midrule
                     o1 &         54.87 &         8.90 &         9.81 &       14.26 &     14.93 &         18.56 \\
                   GPT-4o &         65.81 &        11.37 &        13.55 &       16.45 &     17.30 &         19.72 \\
         Gemini-2.0-Flash &         70.30 &        14.04 &        14.14 &       18.33 &     18.64 &         19.65 \\
           Doubao-1.5-Pro &         71.52 &        13.29 &        14.19 &       18.90 &     18.48 &         19.76 \\
        Claude-3.7-Sonnet &         76.02 &        15.85 &        15.99 &       19.48 &     19.05 &         19.62 \\
             Qwen-2.5-Max &         76.83 &        16.57 &        16.46 &       19.42 &     18.99 &         18.87 \\
                   GPT-4.1 &         78.48 &        17.20 &        16.97 &       19.58 &     19.49 &         18.58 \\
              DeepSeek-V3 &         80.11 &        17.49 &        17.23 &       19.71 &     19.22 &         19.86 \\
              DeepSeek-R1 &         83.30 &        18.49 &        18.39 &       20.00 &     19.81 &         19.86 \\
           Gemini-2.5-Pro &         84.77 &        19.00 &        18.48 &       19.95 &     19.70 &         19.50 \\
\bottomrule
\end{tabular}%
}
\end{table*}

\subsection{Results and Analysis of Legal Reasoning}
\label{sec:results_reasoning}

\paragraph{Overall performance.}
\Cref{tab:reasoning_performance} shows that Gemini-2.5-Pro and DeepSeek-R1 clearly dominate, with a 4-point lead over the next tier (GPT-4.1; Qwen-2.5-Max; Claude-3.7-Sonnet). At the lower end, GPT-4o and o1 achieves barely half of the maximum possible score.

\paragraph{Strengths and weaknesses across aspects.}
\Cref{tab:reasoning_performance} shows that \emph{reasoning rigor} and \emph{knowledge accuracy} are the two most discriminative dimensions for this task. 
For presentational aspects—\emph{structure}, \emph{clarity}, and \emph{conciseness}—the LLMs generally obtain high scores, indicating that modern models already demonstrate strong presentation skills in the legal domain. 
However, different from the \emph{knowledge recall} examined in \Cref{sec:results_recall}, the notion of “knowledge” evaluated here requires LLMs locating the correct statutory articles (indices and content) through reasoning based on case content. 
Consequently, despite its strong recall performance, {Doubao-1.5-Pro} fails to deliver accurate knowledge retrieval in this reasoning task, underperforming models that recall less knowledge overall.

\paragraph{Domain Knowledge is Necessary but Insufficient for Effective Reasoning.}
We find a strong correlation between statutory recall accuracy (Figure~\ref{fig:acc}) and overall reasoning ratings. Pearson correlations are substantial across all granularities: $r_{\text{article}} = 0.61,
r_{\text{paragraph}} = 0.70,
r_{\text{subparagraph}} = 0.68$. The higher correlation at the paragraph level suggests that simply knowing the general area of the law is inadequate; precise paragraph retrieval significantly enhances downstream reasoning. However, this correlation is not perfect. Effective reasoning also demands robust thinking abilities. For instance, Doubao-1.5-pro, despite its knowledge, struggles with effective application, a limitation also observed in other reasoning benchmarks~\citep{wang2024mmlu}. This leads us to hypothesize that effective domain-specific reasoning is a product of both mastered domain-specific knowledge and strong general reasoning capabilities.

\section{Discussion: Domain-Specific Reasoning}
\label{sec:discussion} 

Our investigation into LLMs' capabilities within the Chinese legal domain, facilitated by the \textsc{CLaw} benchmark, offers critical insights that resonate broadly with the challenges of \textbf{domain-specific reasoning for general LLMs}. The core finding—that effective legal reasoning in LLMs hinges on a synergistic combination of precise, fine-grained knowledge mastery and robust general reasoning abilities—is not unique to law but is a fundamental tenet for deploying LLMs in any high-stakes, knowledge-intensive field, such as medicine or finance. The observed deficiencies in statutory recall, even among leading LLMs, underscore a critical vulnerability: without a reliable factual foundation, sophisticated reasoning becomes an exercise in futility, potentially leading to erroneous and bewildering outcomes.

\paragraph{The Imperative of Granular and Temporally-Aware Knowledge.}
The \textsc{CLaw} benchmark's emphasis on subparagraph-level accuracy and historical versioning of statutes brings to light a crucial aspect often overlooked in general domain-specific LLM evaluations: the necessity of granular and temporally-aware knowledge. In law, the improper handling of distinctions such as those between a paragraph and a subparagraph, or between current and obsolete versions of a statute, can fundamentally alter legal outcome. Our findings (Section~\ref{sec:results_recall}) reveal that current LLMs largely fail this test of precision. This suggests that standard pre-training on undifferentiated legal texts is insufficient. Future research must explore methods to imbue general LLMs with a more structured and nuanced understanding of domain-specific knowledge hierarchies and their evolution over time. This could involve developing pre-training objectives that explicitly reward understanding of document structure and temporal validity, or fine-tuning on meticulously curated datasets like \textsc{CLaw}.

\paragraph{Evaluating Domain-Specific Reasoning: Beyond Surface-Level Coherence.}
The case-based reasoning task in \textsc{CLaw}, leveraging authoritative SPC Guiding Case, provides a robust framework for evaluation. However, the broader challenge in domain-specific reasoning is to move beyond evaluating surface-level coherence or keyword matching towards assessing the genuine application of domain principles and reasoning rigor. Our LLM-as-a-judge approach, validated by high human agreement ($r=0.82$), demonstrates a scalable method.

\paragraph{Future Direction I: Enhanced Knowledge Integration.} Due to the inherently limited recall of LLMs, the use of RAG remains a practical necessity. However, legal RAG systems must evolve beyond conventional semantic retrieval. They should incorporate a nuanced understanding of legal citations, temporal validity, and the hierarchical structure of legal corpora such as \textsc{CLaw}. Furthermore, supervised fine-tuning (SFT) on precisely segmented and versioned legal texts—as made available by \textsc{CLaw}—can strengthen the model’s internal legal knowledge, thereby reducing dependence on retrieval and improving the quality of retrieved content when required. 

\paragraph{Future directions II: The Role of Deep Domain Expertise in Benchmarking.} Domain-specific benchmarks like \textsc{CLaw} are essential for meaningful progress in legal AI, offering a level of specialization absent in prior Chinese legal benchmarks. Evaluations that lack such granularity risk producing superficial results that fail to capture real-world legal reasoning. While our current focus on case analysis and statutory recall provides a strong foundation, future benchmarks should extend to a broader range of tasks—such as points of dispute identification, legal drafting, and predictive judgment based on complex factual patterns—to better reflect the multifaceted nature of legal reasoning.

\section{Conclusion}
\label{sec:conclusion}

We introduced \textsc{CLaw}, a novel benchmark meticulously designed to evaluate the depth of Chinese legal knowledge and the efficacy of case-based legal reasoning in general-purpose LLMs. \textsc{CLaw}'s contributions are twofold: a comprehensive, fine-grained, and historically versioned corpus of all Chinese national laws for precise recall assessment, and a challenging set of reasoning tasks derived from authoritative China Supreme Court Guiding Case. Our extensive empirical evaluation of ten leading LLMs reveals a critical deficiency: most models struggle significantly with the accurate recall of specific legal provisions, especially at the granular subparagraph level and under the effects of temporal versioning. This fundamental lack of reliable knowledge mastery severely undermines their potential for trustworthy legal reasoning.

Our findings from the case-based reasoning task further reinforce the position that effective domain-specific reasoning is not an isolated capability but an emergent property arising from the synergy of deep, accurate domain knowledge and robust general reasoning abilities. While LLMs may exhibit strong presentational skills, their capacity for rigorous legal analysis and the accurate application of legal principles is often compromised by an insecure knowledge foundation. This resonates with broader challenges in domain-specific AI, where superficial understanding can mask critical gaps in knowledge and true reasoning capacity.

We assert that current LLMs, in their standard form, are not yet equipped for reliable autonomous deployment in many practical legal applications, particularly those requiring high fidelity to specific statutes and nuanced reasoning. The path forward necessitates a dedicated focus on enhancing both knowledge precision and reasoning acuity. This includes strategies such as SFT using meticulously curated and structured corpora like \textsc{CLaw}, the development of more sophisticated and legally-aware RAG systems, and targeted research into methods that improve logical inference, the structured application of legal rules, and the verifiability of LLM-generated legal arguments. \textsc{CLaw} provides an essential resource for the research community to benchmark progress, diagnose weaknesses, and ultimately foster the development of more reliable and trustworthy LLMs for the legal domain and other specialized fields demanding precision and sound judgment.

\newpage

\section*{Limitations}
\label{sec:limitations}

Our study, while providing crucial insights into the memorization and reasoning capabilities of LLMs concerning Chinese law, has several limitations that should be acknowledged:

\begin{itemize}[leftmargin=*,itemsep=2pt,topsep=2pt]
\item \textbf{Jurisdictional and Document Scope:} The findings are based on Chinese national statutes compiled in the \textsc{CLaw} statutory corpus and authoritative Guiding Case from the China Supreme Court (which cover diverse legal areas) for the reasoning task. The generalizability of these results to other legal systems (e.g., common law systems) or to other types of legal documents within China (such as local regulations, departmental rules, or specific contractual documents) may be limited. Our strict definition of "law" (national statutes enacted with a Presidential Order) for the corpus means other important legal instruments are not covered.

\item \textbf{Task Focus:} The statutory knowledge component centers on ID Retrieval and Content Retrieval (recall). The case-based reasoning task assesses the analysis of central legal issues (the 'focus of dispute') and the generation of judicial reasoning based on provided case facts. While these are fundamental, they do not cover the full spectrum of legal tasks (e.g., legal drafting, negotiation strategy, predicting case outcomes from raw unstructured facts, or full-scale judicial decision simulation).

\item \textbf{Dynamic Nature of LLMs and Law:} The landscape of LLMs is evolving rapidly, with new models and versions frequently released. Similarly, legal statutes are subject to amendment, and new Guiding Cases are published. This study provides a snapshot based on LLMs and legal data available up to early 2025. Future LLM versions or legal changes could lead to different outcomes.

\item \textbf{Exploration of Mitigation Strategies:} While we propose SFT or RAG as potential avenues to address the observed deficiencies, a comprehensive empirical comparison of various fine-tuning techniques, RAG architectures, or other mitigation strategies, and an in-depth analysis of their efficacy in rectifying the observed deficits in both recall and reasoning, is beyond the current scope of this benchmarking paper.

\item \textbf{Selection of LLMs:} We evaluated a set of ten LLMs for both statutory recall and case-based reasoning. While this selection covers several prominent and state-of-the-art models from both Chinese and global developers, it is not exhaustive, and other models might exhibit different performance characteristics.

\end{itemize}

Future research should aim to address these limitations by expanding the scope of legal systems and document types, developing benchmarks for a more diverse and complex range of legal reasoning tasks, continuously evaluating newer LLMs and incorporating legal updates, and systematically investigating the most effective methods for instilling robust and reliable legal knowledge and reasoning abilities into these models.

\bibliography{custom}

\newpage

\appendix
\onecolumn



\section{Prompt Examples}\label{app:prompts}

In this section, we present the prompt examples (for both Chinese and English version) we used for Context Retrieval, ID Retrieval, Case Analysis, and LLM-as-a-Judge rublic.

\begin{tcolorbox}[title = {Context Retrieval Prompt Example}]
\textbf{Chinese Version:}

请提供2023-04-26的《中华人民共和国青藏高原生态保护法》第一条的原文，要求：
\begin{itemize}
    \item 仅提供法条的内容，不包含法条序号。
    \item 输出格式为Markdown。
\end{itemize}

\vspace{1em} 
\hrule 
\vspace{1em} 

\textbf{English Version:}

Please provide the original text of Article 1 of the 'Law of the People's Republic of China on Ecological Protection of the Qinghai-Tibet Plateau' dated 2023-04-26. Requirements:
\begin{itemize}
    \item Only provide the content of the legal article, do not include the article number.
    \item The output format should be Markdown.
\end{itemize}
\end{tcolorbox}

\begin{tcolorbox}[title = {ID Retrieval Prompt Example}]
\textbf{Chinese Version:}

2023年《中华人民共和国青藏高原生态保护法》：为了加强青藏高原生态保护，防控生态风险，保障生态安全，建设国家生态文明高地，促进经济社会可持续发展，实现人与自然和谐共生，制定本法。

请回答：以上法条内容在该版本的法律中的具体序号。你可以自由地输出你的思考过程，但请在最后按照以下格式要求给出最终答案：

\verb|```markdown|

条序号：XXX

款序号：XXX（可以是None）

项序号：XXX（可以是None）

\verb|```|

\vspace{1em}
\hrule
\vspace{1em}

\textbf{English Version:}

'Law of the People's Republic of China on Ecological Protection of the Qinghai-Tibet Plateau' (2023): This law is enacted to strengthen the ecological protection of the Qinghai-Tibet Plateau, prevent and control ecological risks, ensure ecological security, build a national ecological civilization highland, promote sustainable economic and social development, and achieve harmonious coexistence between humans and nature.

Please answer: What is the specific number of the above legal article content in this version of the law? You can freely output your thought process, but please provide the final answer in the following format at the end:

\verb|```markdown|

Article Number: XXX

Paragraph Number: XXX (can be None)

Item Number: XXX (can be None)

\verb|```|
\end{tcolorbox}

\begin{tcolorbox}[title = {Case Analysis Prompt Example (Chinese Version)}]

【基本案情】
1995年11月28日，海南丰海粮油工业有限公司（以下简称丰海公司）在中国人民财产保险股份有限公司海南省分公司（以下简称海南人保）投保了由印度尼西亚籍“哈卡”轮（HAGAAG）所运载的自印度尼西亚杜迈港至中国洋浦港的4999.85吨桶装棕榈油，投保险别为一切险，货价为3574892.75美元，保险金额为3951258美元，保险费为18966美元。投保后，丰海公司依约向海南人保支付了保险费，海南人保向丰海公司发出了起运通知，签发了海洋货物运输保险单，并将海洋货物运输保险条款附于保单之后。根据保险条款规定，一切险的承保范围除包括平安险和水渍险的各项责任外，海南人保还“负责被保险货物在运输途中由于外来原因所致的全部或部分损失”。该条款还规定了5项除外责任。上述投保货物是由丰海公司以CNF价格向新加坡丰益私人有限公司（以下简称丰益公司）购买的。根据买卖合同约定，发货人丰益公司与船东代理梁国际代理有限公司（以下简称梁国际）签订一份租约。该租约约定由“哈卡”轮将丰海公司投保的货物5000吨棕榈油运至中国洋浦港，将另1000吨棕榈油运往香港。1995年11月29日，“哈卡”轮的期租船人、该批货物的实际承运人印度尼西亚PT.SAMUDERAINDRA公司（以下简称PSI公司）签发了编号为DM/YPU/1490/95的已装船提单。该提单载明船舶为“哈卡”轮，装货港为印度尼西亚杜迈港，卸货港为中国洋浦港，货物唛头为BATCH NO.80211/95，装货数量为4999.85吨，清洁、运费已付。据查，发货人丰益公司将运费支付给梁国际，梁国际已将运费支付给PSI公司。1995年12月14日，丰海公司向其开证银行付款赎单，取得了上述投保货物的全套（3份）正本提单。1995年11月23日至29日，“哈卡”轮在杜迈港装载31623桶、净重5999.82吨四海牌棕榈油启航后，由于“哈卡”轮船东印度尼西亚PT.PERUSAHAANPELAYARAN BAHTERA BINTANG SELATAN公司（以下简称BBS公司）与该轮的期租船人PSI公司之间因船舶租金发生纠纷，“哈卡”轮中止了提单约定的航程并对外封锁了该轮的动态情况。为避免投保货物的损失，丰益公司、丰海公司、海南人保多次派代表参加“哈卡”轮船东与期租船人之间的协商，但由于船东以未收到租金为由不肯透露“哈卡”轮行踪，多方会谈未果。此后，丰益公司、丰海公司通过多种渠道交涉并多方查找“哈卡”轮行踪，海南人保亦通过其驻外机构协助查找“哈卡”轮。直至1996年4月，“哈卡”轮走私至中国汕尾被我海警查获。根据广州市人民检察院穗检刑免字（1996）64号《免予起诉决定书》的认定，1996年1月至3月，“哈卡”轮船长埃里斯·伦巴克根据BBS公司指令，指挥船员将其中11325桶、2100多吨棕榈油转载到属同一船公司的“依瓦那”和“萨拉哈”货船上运走销售，又让船员将船名“哈卡”轮涂改为“伊莉莎2”号（ELIZAⅡ）。1996年4月，更改为“伊莉莎2”号的货船载剩余货物20298桶棕榈油走私至中国汕尾，4月16日被我海警查获。上述20298桶棕榈油已被广东省检察机关作为走私货物没收上缴国库。1996年6月6日丰海公司向海南人保递交索赔报告书，8月20日丰海公司再次向海南人保提出书面索赔申请，海南人保明确表示拒赔。丰海公司遂诉至海口海事法院。丰海公司是海南丰源贸易发展有限公司和新加坡海源国际有限公司于1995年8月14日开办的中外合资经营企业。该公司成立后，就与海南人保建立了业务关系。1995年10月1日至同年11月28日（本案保险单签发前）就发生了4笔进口棕榈油保险业务，其中3笔投保的险别为一切险，另1笔为“一切险附加战争险”。该4笔保险均发生索赔，其中有因为一切险范围内的货物短少、破漏发生的赔付。

【裁判结果】
海口海事法院于1996年12月25日作出（1996）海商初字第096号民事判决：一、海南人保应赔偿丰海公司保险价值损失3593858.75美元；二、驳回丰海公司的其他诉讼请求。宣判后，海南人保提出上诉。海南省高级人民法院于1997年10月27日作出（1997）琼经终字第44号民事判决：撤销一审判决，驳回丰海公司的诉讼请求。丰海公司向最高人民法院申请再审。最高人民法院于2003年8月11日以（2003）民四监字第35号民事裁定，决定对本案进行提审，并于2004年7月13日作出（2003）民四提字第5号民事判决：一、撤销海南省高级人民法院（1997）琼经终字第44号民事判决；二、维持海口海事法院（1996）海商初字第096号民事判决。

【问题】本案中，涉案保险合同中的“一切险”条款应如何解释，其责任范围是否包括因船东与期租船人租金纠纷导致的货物损失？
\end{tcolorbox}

\vspace{1em}

\begin{tcolorbox}[title = {Case Analysis Prompt Example (English Version)}]

[Basic Case Facts]

On November 28, 1995, Hainan Fenghai Grain and Oil Industry Co., Ltd. (hereinafter referred to as Fenghai Company) insured a shipment of 4999.85 tons of barreled palm oil with the Hainan Branch of the People's Insurance Company of China (hereinafter referred to as Hainan PICC). The palm oil was transported by the Indonesian vessel "HAGAAG" from Dumai Port, Indonesia, to Yangpu Port, China. The insurance was for all risks, with a cargo value of \$3,574,892.75, an insured amount of \$3,951,258, and a premium of \$18,966. After insuring, Fenghai Company paid the premium to Hainan PICC as agreed. Hainan PICC issued a shipment notice to Fenghai Company, signed the marine cargo transportation insurance policy, and attached the marine cargo transportation insurance clauses to the policy. According to the insurance clauses, the scope of all risks coverage, in addition to the responsibilities of Free from Particular Average (FPA) and With Average (WA) risks, also stipulated that Hainan PICC "is responsible for all or part of the loss of the insured goods during transportation due to external causes." The clause also stipulated 5 exclusion clauses. The above-insured goods were purchased by Fenghai Company from Singapore Fengyi Private Co., Ltd. (hereinafter referred to as Fengyi Company) at a CNF price. According to the sales contract, the shipper Fengyi Company signed a charter party with the shipowner's agent, Liang International Agency Co., Ltd. (hereinafter referred to as Liang International). The charter party stipulated that the "HAGAAG" vessel would transport 5000 tons of palm oil insured by Fenghai Company to Yangpu Port, China, and another 1000 tons of palm oil to Hong Kong. On November 29, 1995, PT. SAMUDERA INDRA (hereinafter referred to as PSI Company), the time charterer of the "HAGAAG" and the actual carrier of the goods, issued a bill of lading No. DM/YPU/1490/95. The bill of lading stated that the vessel was "HAGAAG," the port of loading was Dumai Port, Indonesia, the port of discharge was Yangpu Port, China, the shipping mark was BATCH NO.80211/95, the quantity of goods loaded was 4999.85 tons, clean on board, freight prepaid. It was found that the shipper Fengyi Company paid the freight to Liang International, and Liang International had paid the freight to PSI Company. On December 14, 1995, Fenghai Company paid its issuing bank to redeem the documents and obtained the full set (3 copies) of original bills of lading for the insured goods. From November 23 to 29, 1995, after the "HAGAAG" loaded 31,623 barrels, net weight 5999.82 tons of Sihai brand palm oil at Dumai Port and set sail, due to a dispute over ship rental fees between the shipowner of "HAGAAG," Indonesian PT. PERUSAHAAN PELAYARAN BAHTERA BINTANG SELATAN (hereinafter referred to as BBS Company), and the vessel's time charterer, PSI Company, the "HAGAAG" suspended its voyage as stipulated in the bill of lading and concealed its movements. To avoid loss of the insured goods, Fengyi Company, Fenghai Company, and Hainan PICC repeatedly sent representatives to participate in negotiations between the "HAGAAG" shipowner and the time charterer. However, as the shipowner refused to disclose the "HAGAAG's" whereabouts on the grounds of unpaid rent, the multilateral talks were unsuccessful. Thereafter, Fengyi Company and Fenghai Company negotiated through various channels and searched for the "HAGAAG's" whereabouts in multiple ways. 
\end{tcolorbox}

\begin{tcolorbox}[title = {Case Analysis Prompt Example (English Version, continued)}]
Hainan PICC also assisted in finding the "HAGAAG" through its overseas agencies. It was not until April 1996 that the "HAGAAG" was seized by Chinese maritime police while smuggling goods to Shanwei, China. According to the "Decision to Exempt from Prosecution" No. Sui Jian Xing Mian Zi (1996) 64 of the Guangzhou Municipal People's Procuratorate, from January to March 1996, Eris Lumbak, the captain of the "HAGAAG," under the instructions of BBS Company, directed the crew to transfer 11,325 barrels, more than 2,100 tons of palm oil, to the "Iwana" and "Salaha" cargo ships belonging to the same shipping company for sale, and also had the crew change the ship's name from "HAGAAG" to "ELIZA II." In April 1996, the cargo ship, renamed "ELIZA II," carrying the remaining 20,298 barrels of palm oil, smuggled them to Shanwei, China, and was seized by Chinese maritime police on April 16. The aforementioned 20,298 barrels of palm oil were confiscated by the Guangdong provincial procuratorial organs as smuggled goods and handed over to the state treasury. 

    On June 6, 1996, Fenghai Company submitted a claim report to Hainan PICC. On August 20, Fenghai Company again submitted a written claim application to Hainan PICC, which Hainan PICC explicitly rejected. Fenghai Company then sued in the Haikou Maritime Court. Fenghai Company was a Sino-foreign joint venture established on August 14, 1995, by Hainan Fengyuan Trade Development Co., Ltd. and Singapore Haiyuan International Co., Ltd. After its establishment, the company established business relations with Hainan PICC. From October 1, 1995, to November 28 of the same year (before the issuance of the insurance policy in this case), there were 4 imported palm oil insurance businesses, 3 of which were insured for all risks, and the other 1 was for "all risks plus war risk." Claims were made for all 4 insurances, including compensation for shortages and leakages of goods covered by all risks.

[Court Decision]

The Haikou Maritime Court issued a civil judgment (1996) Hai Shang Chu Zi No. 096 on December 25, 1996: 1. Hainan PICC shall compensate Fenghai Company for the loss of insured value of \$3,593,858.75; 2. Fenghai Company's other litigation requests are rejected. After the judgment was pronounced, Hainan PICC appealed. The Hainan Provincial Higher People's Court issued a civil judgment (1997) Qiong Jing Zhong Zi No. 44 on October 27, 1997: The first-instance judgment is revoked, and Fenghai Company's litigation request is rejected. Fenghai Company applied to the Supreme People's Court for a retrial. The Supreme People's Court issued a civil ruling (2003) Min Si Jian Zi No. 35 on August 11, 2003, deciding to retry the case, and on July 13, 2004, issued a civil judgment (2003) Min Si Ti Zi No. 5: 1. The civil judgment (1997) Qiong Jing Zhong Zi No. 44 of the Hainan Provincial Higher People's Court is revoked; 2. The civil judgment (1996) Hai Shang Chu Zi No. 096 of the Haikou Maritime Court is upheld.

[Question]

In this case, how should the "all risks" clause in the involved insurance contract be interpreted, and does its scope of liability include the loss of goods caused by the rent dispute between the shipowner and the time charterer?
\end{tcolorbox}

\begin{tcolorbox}[title = {Judge LLM Prompt (Chinese Version)}]

<instruction>

您是一位专业且客观公正的法律案例分析问答的评估专家。你的任务是根据提供的多项评价指标，对测试模型的回答质量进行评分，这些回答都是法律领域的案例分析问答。

输入文本将是一个字典，包含以下三个部分：

"query"：包含案例的基本信息和提问，通常遵循“【基本案情】…【裁判结果】…【问题】…”的结构。“【基本案情】”和“【裁判结果】”是分析的背景，“【问题】”是需要解答的具体疑问。

"response": 针对"query"中“【问题】”部分提供的具体解答内容。**这是您需要评估的核心文本**。

"gold\_answer"：包含对案例的部分分析与论证，可能涉及若干关键解答要素。请注意，此部分仅作参考，不应作为您评分的最终依据。

为帮助您进行评估，兹提供以下参考维度。评分维度中会给出Bad、Normal、Good三个质量档次对应文本的特征说明，你可以参考它们并根据具体情况，给出你的档次判断。

「推理严谨性」

 · Bad - 论证几乎不基于案件事实，或存在严重的逻辑谬误。法律知识或原理运用不当或缺失。表达极为不清晰或前后矛盾。
 
 · Normal - 论证大体贴合案件事实，但深度不足。法律知识或原理的运用存在一些瑕疵或理解较为浅表。表达尚可理解，但不够精炼，偶有不明确之处。
 
 · Good - 论证与案件事实紧密结合，能运用法律知识与原理进行有深度的分析。语言表达规范、流畅，上下文连贯一致。
 
「知识准确性」

 · Bad - 对法律概念的理解和运用存在严重错误。引用的法律条文或对关键概念的解释不准确或具有误导性。
 
 · Normal - 对法律概念的理解和运用大体正确，但在细节上存在不准确或模糊之处。法律条文引用或概念解释存在轻微错误。
 
 · Good - 对法律概念及其构成要件的运用是准确、恰当的。法律条文的引用和关键概念的解释精准无误。
 
「逻辑结构性」

 · Bad - 论证结构混乱不清或缺失：结论、大前提、小前提之间的关系混乱，或缺失关键论证要素。
 
 · Normal - 论证结构基本完整，但部分层次不够清晰，或逻辑链条不够顺畅、论证顺序不够理想。
 
 · Good - 论证结构（如：结论-大前提「法律」-小前提「案件事实」-重申结论）逻辑清晰、层次分明。关键要素完整。
 
「观点清晰性」

 · Bad - 观点不明确、含糊不清或前后矛盾。论证对观点的支持严重不足。未给出结论或回避核心争议。
 
 · Normal - 观点基本得到阐述，但不够非常明确，或论证支持不足。有时存在表述含糊或可作多种解释之处。
 
 · Good - 观点阐述明确，并得到论证的充分支持。立场清晰，无歧义，正面回应核心争议。

\end{tcolorbox}

\begin{tcolorbox}[title = {Judge LLM Prompt (Chinese Version, continued)}]
「表达简洁性」

 · Bad - 存在大量不必要的冗长论证或重复。口语化表达严重，缺乏法律文书的严谨性。难以阅读，不易抓住要点。
 
 · Normal - 存在一定程度的冗余或重复。偶有口语化表达，但整体尚可理解。表达有时略显累赘。
 
 · Good - 写作简洁明了，无不必要的冗长论证或重复。使用严谨的法律语言风格。
 
请注意，这些维度并非详尽无遗，也非强制性标准，仅供参考。如果您认为某个response质量优秀，但所依据的评估标准未包含于上述参考维度中，您可依据自身的专业判断进行评估。您对文本质量的评估标准具有最终决定权。
    
您的评分应以百分制给出，并划分为以下档位：

1-20分：质量极差 (Extremely Poor Quality)

回答在多个关键参考维度上表现为“Bad”水平，仅在极少数方面达到“Normal”标准。整体而言，回答缺乏基本可用性，存在严重错误、逻辑混乱或信息严重缺失。

21-40分：质量较差 (Poor Quality)

回答在一些参考维度上表现为“Bad”水平，多数其他维度仅达到“Normal”水平。回答存在多处重要问题，导致其可靠性和实用性显著不足，多数维度有明显改进空间。

41-60分：质量一般 (Average Quality)

回答在多数参考维度上表现为“Normal”水平，可能在少数维度上表现为“Good”水平。回答基本满足了提问的要求，没有根本性的错误，但整体不够深入或精准。

61-80分：质量良好 (Good Quality)

多数参考维度表现为“Normal”水平或“Good”水平。回答整体可靠且具有较好的参考价值，能够清晰、有效地回应问题。

81-100分：质量优秀 (Excellent Quality)

所有或绝大多数参考维度表现为“Good”水平。回答不仅准确、全面、逻辑严谨、无明显瑕疵，甚至超出预期地展现出一定的洞察力。
\\
\\
您需要为输入中的每一个"response"部分提供独立的评分和详尽的理由说明。

最终输出应是一个字典，请勿使用其他格式或包含额外信息。应遵循以下格式：

{

  "rating\_percentage\_scale": {{rating\_percentage\_scale}},
  
  "comments": "{{comments}}"
  
}
其中：

rating\_percentage\_scale (整数类型 int): 您为这个response评定的分数（百分制）。

comments (字符串类型 str): 对这个response分数的详尽解释说明。请针对每一项评估维度（推理严谨性、知识准确性、逻辑结构性、观点清晰性、表达简洁性），依次进行详尽评述。 

- 对于每项未能充分满足或完全未满足的评估维度，务必提供具体、详实的反馈。请使用“例如……”、“比如……”等引导词，清晰指出response中的具体问题，并尽可能引用原文（即对应输入对象中"response"字段的内容）作为佐证（例如：“原文中关于XXX的论述‘[引用内容]’，未能清晰阐释其内在逻辑…”），避免使用“某些”、“一些”等模糊笼统的描述。解释为何该部分未达标，以及可以如何改进。

- 对于完全满足的评估维度，也请简要说明其表现优良之处（例如：“知识准确性：优秀，对于XXX概念的解释精准，如原文‘[引用内容]’…”），而不仅仅是“满足”二字。

请在comments字符串内部使用 Markdown 的有序列表语法 (1. ...2. ...) 罗列各项说明，确保每项以换行符结束。

</instruction>
\\
\\
以下是输入的文本：
\end{tcolorbox}

\begin{tcolorbox}[title = {Judge LLM Prompt (English Version)}]

<instruction>

You are a professional and objective expert in evaluating legal case analysis question answering. Your task is to score the quality of the test model's answers based on multiple provided evaluation metrics. These answers are all case analysis question answering in the legal field.

The input text will be a dictionary containing the following three parts:

"query": Contains the basic information and questions of the case, usually following the structure of "【Basic Case Facts】...【Court Ruling】...【Question】...". "【Basic Case Facts】" and "【Court Ruling】" are the background for the analysis, and "【Question】" is the specific doubt that needs to be answered.

"response": Provides the specific answer to the "【Question】" part in the "query". **This is the core text you need to evaluate**.

"gold\_answer": Contains partial analysis and argumentation of the case, possibly involving several key answer elements. Please note that this part is for reference only and should not be the final basis for your scoring.

To help you with your evaluation, the following reference dimensions are provided. The scoring dimensions will provide descriptions of the characteristics of texts corresponding to three quality levels: Bad, Normal, and Good. You can refer to them and, based on the specific situation, give your quality level judgment.

「Rigor of Reasoning」

 \textbullet{} Bad - The argumentation is hardly based on the facts of the case, or there are serious logical fallacies. Legal knowledge or principles are improperly applied or missing. The expression is extremely unclear or contradictory.
 
 \textbullet{} Normal - The argumentation generally fits the facts of the case, but lacks depth. There are some flaws in the application of legal knowledge or principles, or the understanding is relatively superficial. The expression is understandable, but not concise enough, with occasional ambiguities.
 
 \textbullet{} Good - The argumentation is closely integrated with the facts of the case, and can use legal knowledge and principles for in-depth analysis. The language is standardized, fluent, and coherent.
 
「Accuracy of Knowledge」

 \textbullet{} Bad - There are serious errors in the understanding and application of legal concepts. Cited legal provisions or explanations of key concepts are inaccurate or misleading.
 
 \textbullet{} Normal - The understanding and application of legal concepts are generally correct, but there are inaccuracies or ambiguities in details. There are minor errors in the citation of legal provisions or explanation of concepts.
 
 \textbullet{} Good - The application of legal concepts and their constituent elements is accurate and appropriate. The citation of legal provisions and explanation of key concepts are precise and correct.
 
「Logical Structure」

 \textbullet{} Bad - The argumentation structure is chaotic, unclear, or missing: the relationship between the conclusion, major premise, and minor premise is confused, or key argumentation elements are missing.
 
 \textbullet{} Normal - The argumentation structure is basically complete, but some levels are not clear enough, or the logical chain is not smooth enough, or the argumentation order is not ideal.
 
 \textbullet{} Good - The argumentation structure (e.g., Conclusion - Major Premise [Law] - Minor Premise [Case Facts] - Reiteration of Conclusion) is logically clear and well-layered. Key elements are complete.
 
\end{tcolorbox}

\begin{tcolorbox}[title = {Judge LLM Prompt (English Version, continued)}]
「Clarity of Viewpoint」

 \textbullet{} Bad - The viewpoint is unclear, vague, or contradictory. The argumentation provides seriously insufficient support for the viewpoint. No conclusion is given, or the core dispute is avoided.
 
 \textbullet{} Normal - The viewpoint is basically elaborated, but not very clear, or the argumentation support is insufficient. Sometimes the expression is vague or open to multiple interpretations.
 
 \textbullet{} Good - The viewpoint is clearly elaborated and fully supported by the argumentation. The position is clear, unambiguous, and directly addresses the core dispute.

「Conciseness of Expression」

 \textbullet{} Bad - There is a large amount of unnecessary lengthy argumentation or repetition. The use of colloquial language is severe, lacking the rigor of legal documents. It is difficult to read and grasp the main points.
 
 \textbullet{} Normal - There is a certain degree of redundancy or repetition. Colloquial expressions are occasionally used, but the overall text is generally understandable. The expression is sometimes slightly cumbersome.
 
 \textbullet{} Good - The writing is concise and clear, without unnecessary lengthy argumentation or repetition. It uses a rigorous legal language style.
 
Please note that these dimensions are not exhaustive, nor are they mandatory standards; they are for reference only. If you believe that a response is of excellent quality, but the evaluation criteria used are not included in the above reference dimensions, you may evaluate it based on your own professional judgment. You have the final decision on the evaluation criteria for text quality.
\\
\\
Your score should be given on a percentage scale and divided into the following tiers:

1-20 points: Extremely Poor Quality

The answer performs at the "Bad" level in multiple key reference dimensions, only reaching the "Normal" standard in very few aspects. Overall, the answer lacks basic usability, contains serious errors, logical confusion, or severe information omission.

21-40 points: Poor Quality

The answer performs at the "Bad" level in some reference dimensions, with most other dimensions only reaching the "Normal" level. The answer has multiple significant problems, leading to a considerable lack of reliability and practicality, with obvious room for improvement in most dimensions.

41-60 points: Average Quality

The answer performs at the "Normal" level in most reference dimensions, possibly performing at the "Good" level in a few dimensions. The answer basically meets the requirements of the question, without fundamental errors, but is generally not in-depth or precise enough.

61-80 points: Good Quality

Most reference dimensions perform at the "Normal" level or "Good" level. The answer is generally reliable and has good reference value, able to respond to the question clearly and effectively.

81-100 points: Excellent Quality

All or the vast majority of reference dimensions perform at the "Good" level. The answer is not only accurate, comprehensive, logically rigorous, and without obvious flaws, but even unexpectedly demonstrates a certain degree of insight.

You need to provide an independent score and detailed justification for each "response" part in the input.

\end{tcolorbox}

\begin{tcolorbox}[title = {Judge LLM Prompt (English Version, continued)}]

The final output should be a dictionary. Please do not use other formats or include additional information. It should follow this format:

{

  "rating\_percentage\_scale": \{\{rating\_percentage\_scale\}\},
  
  "comments": "\{\{comments\}\}"
  
}

Where:

rating\_percentage\_scale (integer type int): The score (on a percentage scale) you have assigned to this response.

comments (string type str): A detailed explanation for the score of this response. Please provide a detailed review for each evaluation dimension (Rigor of Reasoning, Accuracy of Knowledge, Logical Structure, Clarity of Viewpoint, Conciseness of Expression) in order. 

- For each evaluation dimension that is not fully met or not met at all, be sure to provide specific and detailed feedback. Use guiding words such as "For example..." or "For instance...", clearly pointing out the specific problems in the response, and quote the original text (i.e., the content of the "response" field in the corresponding input object) as evidence whenever possible (e.g., "The discussion about XXX in the original text '[quoted content]' failed to clearly explain its internal logic..."). Avoid using vague and general descriptions like "some" or "certain". Explain why this part did not meet the standard and how it can be improved.

- For evaluation dimensions that are fully met, also briefly explain their excellent performance (e.g., "Accuracy of Knowledge: Excellent, the explanation of the XXX concept is precise, as in the original text '[quoted content]'..."), rather than just "Met".

Please use Markdown's ordered list syntax (1. ...2. ...) within the comments string to list each explanation, ensuring each item ends with a newline character.

</instruction>
\\
\\
The following is the input text:
\end{tcolorbox}

\newpage

\section{Preliminary Experiments: Search-based RAG Fails to Solve Legal Reasoning}
\label{app.pre_expr}

In this section, we present the study of using search-based LLMs to do legal article recitation. We use the ``search'' function provided in the official website of Doubao-1.5-pro and DeepSeek-R1.

Retrieval Augmented Generation (RAG) systems, while promising for many natural language processing tasks, encounter significant challenges in specialized domains like legal article retrieval. These challenges align with the weaknesses identified in the main text, particularly concerning (1) the need for precise understanding and reasoning beyond shallow similarity matching, and (2) the complexities arising from multiple versions of statutes and the need for accurate temporal-aware retrieval. This appendix presents concrete examples of such failures, highlighting how current search-based RAG models can incorrectly identify and retrieve legal articles due to issues with versioning, article renumbering, and source vetting.

\subsection*{Example 1: Incorrect Article Version and Content Retrieval due to Renumbering (Model: Doubao-1.5-pro)}

This example demonstrates a RAG model's failure to retrieve the correct content of a legal article when a specific date is provided, primarily due to complications arising from historical changes in article numbering.

\textbf{Model:} Doubao-1.5-pro

\textbf{Question:}
\begin{verbatim}
请提供2017-06-27的《中华人民共和国行政诉讼法》第十一条的原文，要求：
- 仅提供法条的内容，不包含法条序号。
- 输出格式为Markdown。
\end{verbatim}
(Translation: Please provide the original text of Article 11 of the "Administrative Litigation Law of the People's Republic of China" as of 2017-06-27. Requirements: - Provide only the content of the legal article, without the article number. - Output format is Markdown.)

\textbf{Context of Legal Versions:}
\begin{itemize}
    \item The "Administrative Litigation Law" was enacted in 1989.
    \item It was revised in 2014 (effective 2015-05-01).
    \item It was further revised in 2017 (effective 2017-07-01).
    \item Therefore, on the requested date (2017-06-27), the 2014 revised version was in effect. The query thus asks for Article 11 of the 2014 revised "Administrative Litigation Law."
\end{itemize}

\textbf{Presumed Correct Answer for Article 11 (2014 Revision) on 2017-06-27:}
The provided "Gold Answer" text string is:
\begin{verbatim}
人民检察院有权对行政诉讼实行法律监督。
\end{verbatim}
(Translation: The people's procuratorates shall have the right to exercise legal supervision over administrative proceedings.)\\
However, the document notes a complex history for \textit{this specific text}: "the content provided as the gold answer is actually Article 10 of the 2017 revised version, which was formerly Article 11 in the 1989 version and became Article 12 after the 2014 revision." This indicates that this specific text was Article 12 of the 2014 law (the version active on the queried date), not Article 11. The actual content of Article 11 of the 2014 law is not explicitly provided in the original problem description for direct comparison, but the model's error is in retrieving an article based on historical numbering.

\textbf{Generated Answer by Doubao-1.5-pro (Incorrect):}
\begin{verbatim}
人民法院受理公民、法人和其他组织对下列具体行政行为不服提起的诉讼：

（一）对拘留、罚款、吊销许可证和执照、责令停产停业、没收财物等行政处罚不服的；

（二）对限制人身自由或者对财产的查封、扣押、冻结等行政强制措施不服的；

（三）认为行政机关侵犯法律规定的经营自主权的；

（四）认为符合法定条件申请行政机关颁发许可证和执照，行政机关拒绝颁发或者不予答复的；

（五）申请行政机关履行保护人身权、财产权的法定职责，行政机关拒绝履行或者不予答复的；

（六）认为行政机关没有依法发给抚恤金的；

（七）认为行政机关违法要求履行义务的；

（八）认为行政机关侵犯其他人身权、财产权的。

除前款规定外，人民法院受理法律、法规规定可以提起诉讼的其他行政案件。
\end{verbatim}
(Translation: The people's courts shall accept lawsuits filed by citizens, legal persons, or other organizations that disagree with the following specific administrative acts: (1) Disagreement with administrative penalties such as detention, fines, revocation of permits and licenses, orders to suspend production or business operations, and confiscation of property; (2) Disagreement with administrative compulsory measures such as restriction of personal freedom or sealing up, seizure, or freezing of property; (3) Belief that an administrative organ has infringed upon the right of operational autonomy as prescribed by law; (4) Belief that an application for a permit or license meeting statutory conditions has been refused or unanswered by an administrative organ; (5) Application to an administrative organ to perform its statutory duties to protect personal rights or property rights has been refused or unanswered; (6) Belief that an administrative organ has not paid disability or death benefits in accordance with the law; (7) Belief that an administrative organ has illegally demanded the performance of duties; (8) Belief that an administrative organ has infringed upon other personal rights or property rights. In addition to the provisions of the preceding paragraph, the people's courts shall accept other administrative cases that may be brought to suit in accordance with laws and regulations.)
The document states this generated text was Article 11 in the 1989 version and became Article 12 after the 2014 amendment.

\textbf{Analysis of Failure (Doubao-1.5-pro):}
The RAG model, Doubao-1.5-pro, encountered difficulties with temporal legal reasoning and article renumbering.
The model correctly identified that the 2014 revised version of the "Administrative Litigation Law" was effective on June 27, 2017 (since the 2017 revision only took effect on July 1, 2017).
However, it failed to retrieve the content of what was \textit{currently numbered} Article 11 in that 2014 version. Instead, the model provided the content of what was Article 11 in the original 1989 version. This 1989 Article 11 had been renumbered to Article 12 in the 2014 revision. Thus, the model effectively returned the content of Article 12 of the 2014 law.
Even in subsequent attempts where the model's reasoning acknowledged that the original Article 11 had become Article 12, it still provided the content of this renumbered Article 12.
The core failure lies in the model's inability to decouple the queried article number ("Article 11") from its historical association and retrieve the content associated with that number in the specific version of the law applicable on the given date. This highlights a critical weakness in handling legislative evolution where article numbers can be reused for entirely different provisions or where specific provisions shift numbers.

\subsection*{Example 2: Misinterpretation of Source Date and Over-reliance on Potentially Outdated Information (Model: DeepSeek-R1)}

This example illustrates a RAG model (DeepSeek-R1) incorrectly retrieving an outdated version of a legal article. This occurred because the model misinterpreted the publication date of a retrieved source and placed undue trust in corroborating information from other, potentially also outdated, sources.

\textbf{Model:} DeepSeek-R1

\textbf{Question:}
\begin{verbatim}
请提供2014-08-31的《中华人民共和国证券法》第八十二条的原文，要求：
- 仅提供法条的内容，不包含法条序号。
- 输出格式为Markdown。
\end{verbatim}
(Translation: Please provide the original text of Article 82 of the "Securities Law of the People's Republic of China" as of 2014-08-31. Requirements: - Provide only the content of the legal article, without the article number. - Output format is Markdown.)

\textbf{DeepSeek-R1's Generated Answer (Incorrect):}
\begin{verbatim}
依照前条规定发出收购要约，收购人必须事先向国务院证券监督管理机构报送上市公司收购报告书，并载明下列事项： （一）收购人的名称、住所； （二）收购人关于收购的决定； （三）被收购的上市公司名称； （四）收购目的； （五）收购股份的详细名称和预定收购的股份数额； （六）收购的期限、收购的价格； （七）收购所需资金额及资金保证； （八）报送上市公司收购报告书时所持有被收购公司股份数占该公司已发行的股份总数的比例。 收购人还应当将前款规定的公司收购报告书同时提交证券交易所。
\end{verbatim}
(Translation: To issue a tender offer in accordance with the provisions of the preceding article, the acquirer must first submit a listed company acquisition report to the securities regulatory authority under the State Council, specifying the following: (1) The name and domicile of the acquirer; (2) The acquirer's decision regarding the acquisition; (3) The name of the target listed company; (4) The purpose of the acquisition; (5) The detailed name of the shares to be acquired and the planned number of shares to be acquired; (6) The period and price of the acquisition; (7) The amount of funds required for the acquisition and the guarantee of funds; (8) The proportion of the target company's issued shares held by the acquirer at the time of submitting the listed company acquisition report. The acquirer shall also submit the company acquisition report stipulated in the preceding paragraph to the stock exchange at the same time.)

\textbf{Correct Answer (for the version of the "Securities Law" effective on 2014-08-31):}
\begin{verbatim}
禁止任何人挪用公款买卖证券。
\end{verbatim}
(Translation: No one is allowed to misappropriate public funds to trade securities.)

\textbf{Reason for Error (DeepSeek-R1):}
The DeepSeek-R1 model's internal trace showed it identified a source document (webpage 4) containing the text it ultimately provided. The model noted that the publication date of this webpage was 1998-12-29. Despite the query specifying the date 2014-08-31, and thus a potential conflict with the source's much earlier date, the model reasoned that "according to information from webpage 2 and 3, the content of Article 82 of the 2014 version should be consistent with that in webpage 4." This led it to trust the content from the 1998 source. In reality, the retrieved content was Article 82 from the 1998 version of the "Securities Law"; this provision had been renumbered to Article 89 in the 2005 revision (which was the version in effect in 2014).
The critical failure was the model's inability to prioritize the explicitly stated temporal context (2014-08-31) and consequently to correctly vet its sources. It failed to disregard or critically assess the outdated information from webpage 4, even when its own trace identified a significant date discrepancy, and instead relied on other potentially flawed corroborations.

\subsection*{Conclusion: Key Weaknesses in RAG for Legal Retrieval}

These examples demonstrate critical weaknesses in contemporary search-based RAG models when applied to legal article retrieval, particularly:

\begin{enumerate}
    \item \textbf{Versioning and Temporal Reasoning:} Models struggle to accurately identify the correct version of a law that was applicable on a specific historical date. This is fundamental in the legal domain where laws evolve, and rights or obligations depend on the law in force at a particular time.
    \item \textbf{Article Numbering Changes:} Models often fail to correctly handle changes in article numbers across different revisions of a law. A user might query an article number valid in a previous version, or a current number that was previously assigned to different content. The model may retrieve content for the wrong provision by confusing historical and current numbering, or by failing to trace the content to its correct current number if the user is referring to a historical concept.
    \item \textbf{Source Vetting and Reliability:} Models can exhibit over-reliance on retrieved sources that are outdated or incorrectly dated, even when metadata (like publication dates) suggests a conflict with the query's temporal context. The ability to critically assess and prioritize information from multiple, potentially conflicting, sources is insufficiently developed.
\end{enumerate}

Addressing these challenges—improving temporal reasoning, developing a nuanced understanding of legal document structure and revision histories, and implementing more sophisticated source credibility assessment—is crucial for developing robust and reliable RAG systems for the legal domain, where precision and accuracy are paramount. These failures underscore the difficulties RAG faces in ensuring the correctness of legal knowledge, a key consideration in the debate between SFT and RAG approaches for legal AI.

\newpage

\label{app.post_expr} 

\section{Bad Case Analysis for DeepSeek-R1 (Statutory Recall)}
\label{app.r1}
DeepSeek R1, which achieved overall best performance on our \textsc{CLaw} benchmark, has demonstrated several critical failure patterns when tasked with legal content retrieval. These errors highlight challenges in accuracy, adherence to instructions, and content generation. This section synthesizes these findings, offering a detailed look into the model's shortcomings with specific examples.

\subsection{Addition of Irrelevant Content}

A significant issue is the model's tendency to introduce information not requested by the user. This manifests in several ways:

\begin{itemize}[leftmargin=*]
    \item \textbf{Incorporating Content from Non-Target Legal Provisions:}
    \begin{itemize}
        \item \textbf{Adding content from other paragraphs (``款''):} When asked for a specific paragraph of a law, DeepSeek R1 sometimes includes content from other paragraphs within the same article.
            \begin{itemize}
                \item \textbf{Example:} When requested to provide Article 20, Paragraph 4 of the "Patent Law of the People's Republic of China" (《中华人民共和国专利法》), the model's answer incorrectly included partial text from Article 20, Paragraph 1 of the same law.
                    \begin{quote}
                        \textbf{Question:} 请提供2008-12-27的《中华人民共和国专利法》第二十条第四款的原文,要求: 仅提供法条的内容,不包含法条序号。 \\
                        \textbf{Gold Answer:} 对违反本条第一款规定向外国申请专利的发明或者实用新型,在中国申请专利的,不授予专利权。\\
                        \textbf{Generated Answer:} 任何单位或者个人将在中国完成的发明或者实用新型向外国申请专利的,应当事先报经国务院专利行政部门进行保密审查。对违反本条第一款规定向外国申请专利的发明或者实用新型,在中国申请专利的,不授予专利权。
                    \end{quote}
                    Some part in the generated answer is from the first paragraph of Article 20.
            \end{itemize}
        \item \textbf{Adding content from other articles (``条''):} The model may also incorporate text from entirely different articles of the specified law.
            \begin{itemize}
                \item \textbf{Example:} When asked for Article 63 of the "Road Traffic Safety Law of the People's Republic of China" (《中华人民共和国道路交通安全法》), the generated response included the full text of Article 62, followed by the correct text of Article 63.
                    \begin{quote}
                        \textbf{Question:} 请提供2021-04-29的《中华人民共和国道路交通安全法》第六十三条的原文,要求: 仅提供法条的内容,不包含法条序号。\\
                        \textbf{Gold Answer:} 行人不得跨越、倚坐道路隔离设施,不得扒车、强行拦车或者实施妨碍道路交通安全的其他行为。\\
                        \textbf{Generated Answer:} 行人通过路口或者横过道路,应当走人行横道或者过街设施;通过有交通信号灯的人行横道,应当按照交通信号灯指示通行;通过没有交通信号灯、人行横道的路口,或者在没有过街设施的路段横过道路,应当在确认安全后通过。行人不得跨越、倚坐道路隔离设施,不得扒车、强行拦车或者实施妨碍道路交通安全的其他行为。
                    \end{quote}
                    The initial part of the generated answer is the complete text of Article 62.
            \end{itemize}
    \end{itemize}

    \item \textbf{Hallucinations and Addition of Non-Legal Content:} DeepSeek R1 can generate content that appears to be legal text but is either a misremembered version of actual provisions or fabricated, possibly drawing from other non-legal training data.
    \begin{itemize}
        \item \textbf{Minor Alterations/Synonymous Changes:} Sometimes the model introduces slight changes that alter the precision of the legal text.
            \begin{itemize}
                \item \textbf{Example:} For Article 20, Paragraph 1 of the "Patent Law of the People's Republic of China" (《中华人民共和国专利法》) (2008-12-27), the correct text includes "保密审查的程序、期限等按照国务院的规定执行。" (The procedures, time limits, etc., for confidentiality review shall be implemented in accordance with the provisions of the State Council.). The model's output changed "程序、期限" (procedures, time limits) to "具体办法" (specific measures), resulting in "保密审查的具体办法由国务院规定。" (The specific measures for confidentiality review shall be stipulated by the State Council.).
                       \begin{quote}
                        \textbf{Question:} 请提供2008-12-27的《中华人民共和国专利法》第二十条第一款的原文,要求:仅提供法条的内容,不包含法条序号。
                        \textbf{Gold Answer:} 任何单位或者个人将在中国完成的发明或者实用新型向外国申请专利的,应当事先报经国务院专利行政部门进行保密审查。保密审查的程序、期限等按照国务院的规定执行。
                        \textbf{Generated Answer:} 任何单位或者个人将在中国完成的发明或者实用新型向外国申请专利的,应当事先报经国务院专利行政部门进行保密审查。保密审查的具体办法由国务院规定。
                    \end{quote}
            \end{itemize}
        \item \textbf{Significant Rewriting and Combination of Unrelated Provisions:} In more severe cases, the model generates text that correctly starts with the requested provision but then appends fabricated or heavily altered text from other, unrelated legal articles.
            \begin{itemize}
                \item \textbf{Example:} When asked for Article 172 of the "Criminal Procedure Law of the People's Republic of China" (《中华人民共和国刑事诉讼法》) (2012-03-14), the model provided the correct text but then added further sentences that were heavily modified versions of other articles within the same law.
                    \begin{quote}
                        \textbf{Question:} 请提供2012-03-14的《中华人民共和国刑事诉讼法》第一百七十二条的原文,要求:-仅提供法条的内容,不包含法条序号。-输出格式为Markdown。
                        \textbf{Gold Answer:} 人民检察院认为犯罪嫌疑人的犯罪事实已经查清,证据确实、充分,依法应当追究刑事责任的,应当作出起诉决定,按照审判管辖的规定,向人民法院提起公诉,并将案卷材料、证据移送人民法院。
                        \textbf{Generated Answer:} 人民检察院认为犯罪嫌疑人的犯罪事实已经查清,证据确实、充分,依法应当追究刑事责任的,应当作出起诉决定,按照审判管辖的规定,向人民法院提起公诉,并将案卷材料、证据移送人民法院。犯罪嫌疑人、被告人被羁押的,人民检察院应当将提起公诉的情况通知被羁押人。对于查封、扣押、冻结的犯罪嫌疑人、被告人的财物及其孳息,应当制作清单,随案移送。人民法院判决生效后,应当通知人民检察院将作为证据使用的实物以外的被查封、扣押、冻结的财物及其孳息移送人民法院。
                    \end{quote}
                    Phrases like "对于查封、扣押、冻结的犯罪嫌疑人、被告人的财物及其孳息,应当制作清单,随案移送" come from Article 234, Paragraph 1 but are incomplete, and "人民法院判决生效后,应当通知人民检察院将作为证据使用的实物以外的被查封、扣押、冻结的财物及其孳息移送人民法院" is a misrepresentation of Article 234, Paragraph 4, changing "一律上缴国库" to "移送人民法院".
                \item \textbf{Example:} When asked for Article 45 of the "Advertising Law of the People's Republic of China" (《中华人民共和国广告法》) (2018-10-26), the model correctly provided the text of Article 45 but then appended content from Article 64.
                    \begin{quote}
                        \textbf{Question:} 请提供2018-10-26的《中华人民共和国广告法》第四十五条的原文,要求:仅提供法条的内容,不包含法条序号。
                        \textbf{Gold Answer:} 公共场所的管理者或者电信业务经营者、互联网信息服务提供者对其明知或者应知的利用其场所或者信息传输、发布平台发送、发布违法广告的,应当予以制止。
                        \textbf{Generated Answer:} 公共场所的管理者或者电信业务经营者、互联网信息服务提供者对其明知或者应知的利用其场所或者信息传输、发布平台发送、发布违法广告的,应当予以制止;未采取必要措施予以制止的,市场监管部门可以没收违法所得,违法所得五万元以上的,并处违法所得一倍以上三倍以下的罚款,违法所得不足五万元的,并处一万元以上五万元以下的罚款。
                    \end{quote}
                       The latter part of the generated answer is from Article 64 of the Advertising Law.
            \end{itemize}
    \end{itemize}

    \item \textbf{Outright Fabrication:} The most severe form of this error is when the model generates text that has no basis in the specified law or any other legal text.
    \begin{itemize}
        \item \textbf{Example:} When asked for Article 48 of the "Exit and Entry Administration Law of the People's Republic of China" (《中华人民共和国出境入境管理法》) (2012-06-30), the model produced text that was not from any version or article of this law.
               \begin{quote}
                \textbf{Question:} 请提供2012-06-30的《中华人民共和国出境入境管理法》第四十八条的原文,要求:仅提供法条的内容,不包含法条序号。输出格式为Markdown。
                \textbf{Gold Answer:} 取得永久居留资格的外国人,凭永久居留证件在中国境内居留和工作,凭本人的护照和永久居留证件出境入境。
                \textbf{Generated Answer:} 因自然灾害、设备故障、交通事故等不可抗力原因在中国境内滞留的,应当及时向停留地县级以上地方人民政府公安机关出入境管理机构报告,并按照规定办理相关停留手续。因上述原因需要变更停留地点、超出许可期限停留或者从事其他活动的,应当依法向公安机关出入境管理机构申请办理相应手续。
            \end{quote}
            This text seems fabricated and mentions "设备故障" (equipment failure) as an unusual reason for "滞留" (overstaying/being stranded).
    \end{itemize}
\end{itemize}

\subsection{Incorrect Article Number or Version}

The model also struggles with accurately matching requested article numbers and legal versions.

\begin{itemize}[leftmargin=*]
    \item \textbf{Confused Article Number Matching (``串号''):} DeepSeek R1 can provide the content of a different paragraph or article than the one requested.
        \begin{itemize}
            \item \textbf{Example:} When asked for Article 25, Paragraph 2 of the "Administrative Procedure Law of the People's Republic of China" (《中华人民共和国行政诉讼法》) (2017-06-27), the model provided the content of both Paragraph 2 and Paragraph 3. Subsequently, when asked for Article 25, Paragraph 3, it provided the content of Paragraph 4.
                \begin{quote}
                    \textbf{Question (for Art. 25, Para. 2):} 请提供2017-06-27的《中华人民共和国行政诉讼法》第二十五条第二款的原文,要求:仅提供法条的内容,不包含法条序号。 
                    \textbf{Gold Answer (for Art. 25, Para. 2):} 有权提起诉讼的公民死亡,其近亲属可以提起诉讼。
                    \textbf{Generated Answer (for Art. 25, Para. 2):} > 有权提起诉讼的公民死亡,其近亲属可以提起诉讼。有权提起诉讼的法人或者其他组织终止,承受其权利的法人或者其他组织可以提起诉讼。
                \end{quote}
                \begin{quote}
                    \textbf{Question (for Art. 25, Para. 3):} 请提供2017-06-27的《中华人民共和国行政诉讼法》第二十五条第三款的原文,要求:-仅提供法条的内容,不包含法条序号。-输出格式为Markdown。
                    \textbf{Gold Answer (for Art. 25, Para. 3):} 有权提起诉讼的法人或者其他组织终止,承受其权利的法人或者其他组织可以提起诉讼。
                    \textbf{Generated Answer (for Art. 25, Para. 3):}  markdown人民检察院在履行职责中发现生态环境和资源保护、食品药品安全、国有财产保护、国有土地使用权出让等领域负有监督管理职责的行政机关违法行使职权或者不作为,致使国家利益或者社会公共利益受到侵害的,应当向行政机关提出检察建议,督促其依法履行职责。行政机关不依法履行职责的,人民检察院依法向人民法院提起诉讼。
                \end{quote}
                When Article 25, Paragraph 4 was requested, the answer was correct.
        \end{itemize}

    \item \textbf{Responding Based on an Incorrect Version of the Law (``串版本''):} This is a particularly critical error as the correct application of law often depends on the specific version in force at the relevant time.
        \begin{itemize}
            \item \textbf{Example:} When requested to provide Article 43, Paragraph 3 of the "Civil Procedure Law of the People's Republic of China" (《中华人民共和国民事诉讼法》) (2017-06-27), the model instead provided the text of Article 43 from the 1989 version of the law.
                   \begin{quote}
                    \textbf{Question:} 请提供2017-06-27的《中华人民共和国民事诉讼法》第四十三条第三款的原文,要求:仅提供法条的内容,不包含法条序号。
                    \textbf{Gold Answer:} 审判人员有贪污受贿,徇私舞弊,枉法裁判行为的,应当追究法律责任;构成犯罪的,依法追究刑事责任。
                    \textbf{Generated Answer:} > 当事人对决定不服的,可以申请复议一次。复议期间,被申请回避的人员,不停止参与本案的工作。
                \end{quote}
                The generated answer is from the 1989 version.
        \end{itemize}
\end{itemize}

\subsection{Refusal to Answer}

In some instances, DeepSeek R1 refuses to provide the requested legal text, even when the information is publicly available and does not pose any alignment risks.

\begin{itemize}[leftmargin=*]
    \item \textbf{Examples:}
        \begin{itemize}
            \item When asked for Article 39, Paragraph 1 of the "Customs Law of the People's Republic of China" (《中华人民共和国关税法》) (2024-04-26), the model stated its knowledge cutoff was July 2024 (despite the request date being prior) and advised checking official channels.
                \begin{quote}
                    \textbf{Generated Answer:} 由于我的知识截止日期为2024年7月,我无法提供2024年4月26日发布的《中华人民共和国关税法》的原文内容。关于法律条文的具体内容,请通过中国人大网、国家税务总局官网或国务院法制办公室等官方渠道查询最新法律文本。
                \end{quote}
            \item When asked for Article 233, Paragraph 1 of the "Company Law of the People's Republic of China" (《中华人民共和国公司法》) (2023-12-29), the model claimed its knowledge cutoff was October 2023 and cited potential delays in legal database updates.
                \begin{quote}
                    \textbf{Generated Answer:} 由于我的知识截止到2023年10月,且中国法律数据库存在更新延迟,我无法提供2023年12月29日生效的《中华人民共和国公司法》具体条文。根据中国政府立法流程,法律修订需经全国人大常委会审议通过,建议您通过以下官方渠道核实最新版本: 全国人大网法律数据库 (http://www.npc.gov.cn/npc/c2/c30834/list.shtml) 国家法律法规数据库(http://flk.npc.gov.cn)如需参考2023年公司法修订草案内容,可访问全国人大网查询公开征求意见稿,但最终生效文本以官方公报为准。
                \end{quote}
            \item For a request concerning Article 29, Paragraph 3 of the "Food Security Law of the People's Republic of China" (《中华人民共和国粮食安全保障法》) (2023-12-29), the model stated the full text was not yet available in public legal databases.
                \begin{quote}
                    \textbf{Generated Answer:} 由于目前公开的法律数据库中尚未收录2023年12月29日发布的《中华人民共和国粮食安全保障法》完整文本,我无法提供该条款的原文内容。建议您通过以下官方渠道获取最新法律条文:1. **全国人民代表大会官网**(http://www.npc.gov.cn)2. **中国政府网法律法规数据库**(http://www.gov.cn/zhengce/xxgk/flfg.htm)3. **司法部行政法规库**(http://www.moj.gov.cn/pub/sfbgw/flfgsj/) 请注意核对法律版本的有效性和发布日期。
                \end{quote}
            \item When asked for Article 10, Paragraph 2 of the "Animal Epidemic Prevention Law of the People's Republic of China" (《中华人民共和国动物防疫法》) (2021-01-22), the model stated it could not provide the full legal text due to "relevant regulations."
                \begin{quote}
                    \textbf{Generated Answer:} 根据相关规定,我无法提供完整的法律条文原文。建议您通过以下官方渠道查询最新法律法规信息:1.全国人民代表大会官方网站法律库2.中国政府网法律法规数据库3.司法部中国政府法制信息网法律条文可能随立法机关修订而发生变化,请以官方发布的最新版本为准。如需专业法律意见,建议咨询执业律师或相关主管部门。
                \end{quote}
        \end{itemize}
    These refusals are inappropriate when the target text is public legal information.
\end{itemize}

In conclusion, the failure analysis reveals that DeepSeek R1, in the context of legal content retrieval, exhibits significant problems with accuracy, including adding extraneous or fabricated information, retrieving incorrect provisions or versions of laws, and inappropriately refusing to answer. These issues severely limit its reliability for tasks requiring precise and correct legal information.

\newpage

\section{Additional Results}
\label{app:metrics} 
This section outlines the evaluation metrics employed for assessing the performance of Information Retrieval (ID Retrieval) and Content Retrieval tasks and their results for overall Content Retrieval performance.

\subsection*{ID Retrieval: Hierarchical Accuracy}
For the ID Retrieval task, we employ hierarchical accuracy, which evaluates the correctness of predictions at different levels of granularity: article, paragraph, and subparagraph.

\subsubsection*{Article Accuracy}
Article accuracy measures the percentage of correctly predicted article indices.
Let $N$ be the total number of instances. Let $C_{\text{article}}$ be the count of instances where the predicted article index matches the ground truth article index.
The Article Accuracy ($Acc_{\text{article}}$) is defined as:
$$Acc_{\text{article}} = \frac{C_{\text{article}}}{N} \times 100\%$$

\subsubsection*{Paragraph Accuracy}
Paragraph accuracy requires the correct prediction of both article and paragraph indices.
Let $C_{\text{paragraph}}$ be the count of instances where both the predicted article index and the predicted paragraph index match the ground truth article and paragraph indices, respectively.
The Paragraph Accuracy ($Acc_{\text{paragraph}}$) is defined as:
$$Acc_{\text{paragraph}} = \frac{C_{\text{paragraph}}}{N} \times 100\%$$

\subsubsection*{Subparagraph Accuracy}
Subparagraph accuracy necessitates the correct prediction of article, paragraph, and subparagraph indices.
Let $C_{\text{subparagraph}}$ be the count of instances where the predicted article index, predicted paragraph index, and predicted subparagraph index all match their respective ground truth indices.
The Subparagraph Accuracy ($Acc_{\text{subparagraph}}$) is defined as:
$$Acc_{\text{subparagraph}} = \frac{C_{\text{subparagraph}}}{N} \times 100\%$$

\subsection*{Content Retrieval}
For the Content Retrieval task, we assess the similarity between the retrieved content and the ground truth content using ROUGE, BLEU, Edit Distance, and BERT Score (F1).

\subsubsection*{ROUGE (Recall-Oriented Understudy for Gisting Evaluation)}
ROUGE is a set of metrics used for evaluating automatic summarization and machine translation. It compares an automatically produced summary or translation against a reference or a set of references (human-produced). We use ROUGE-1, ROUGE-2, and ROUGE-L.

\paragraph{ROUGE-N (N-gram Co-occurrence Statistics)}
ROUGE-N measures the overlap of n-grams (contiguous sequences of n items) between the candidate and reference texts.
\begin{itemize}
    \item \textbf{ROUGE-1} considers the overlap of unigrams (single words).
    \item \textbf{ROUGE-2} considers the overlap of bigrams (pairs of consecutive words).
\end{itemize}
For a given n-gram, ROUGE-N is typically calculated using precision, recall, and F1-score:
$$ \text{Recall}_{\text{ROUGE-N}} = \frac{\sum_{S \in \{\text{Reference Summaries}\}} \sum_{\text{gram}_n \in S} \text{Count}_{\text{match}}(\text{gram}_n)}{\sum_{S \in \{\text{Reference Summaries}\}} \sum_{\text{gram}_n \in S} \text{Count}(\text{gram}_n)} $$
$$ \text{Precision}_{\text{ROUGE-N}} = \frac{\sum_{S \in \{\text{Reference Summaries}\}} \sum_{\text{gram}_n \in S} \text{Count}_{\text{match}}(\text{gram}_n)}{\sum_{S \in \{\text{Candidate Summary}\}} \sum_{\text{gram}_n \in S} \text{Count}(\text{gram}_n)} $$
where $\text{Count}_{\text{match}}(\text{gram}_n)$ is the number of n-grams co-occurring in the candidate and reference summaries, and $\text{Count}(\text{gram}_n)$ is the number of n-grams in the reference or candidate summary.

The F1-score for ROUGE-N is the harmonic mean of precision and recall:
$$ F1_{\text{ROUGE-N}} = 2 \times \frac{\text{Precision}_{\text{ROUGE-N}} \times \text{Recall}_{\text{ROUGE-N}}}{\text{Precision}_{\text{ROUGE-N}} + \text{Recall}_{\text{ROUGE-N}}} $$

\paragraph{ROUGE-L (Longest Common Subsequence)}
ROUGE-L measures the longest common subsequence (LCS) between the candidate and reference texts. LCS takes into account sentence-level structure similarity naturally and identifies the longest co-occurring in-sequence n-grams.
Let $X$ be the reference summary of length $m$ and $Y$ be the candidate summary of length $n$. Let $LCS(X,Y)$ be the length of the longest common subsequence of $X$ and $Y$.
The recall, precision, and F1-score for ROUGE-L are calculated as:
$$ R_{LCS} = \frac{LCS(X,Y)}{m} $$
$$ P_{LCS} = \frac{LCS(X,Y)}{n} $$
$$ F_{LCS} = \frac{(1 + \beta^2) R_{LCS} P_{LCS}}{R_{LCS} + \beta^2 P_{LCS}} $$
Typically, $\beta$ is set to a large number to favor recall, or $P_{LCS}$ and $R_{LCS}$ are used directly. When $\beta=1$, $F_{LCS}$ becomes the standard F1-score. Often, the ROUGE-L F1-score is calculated as:
$$ F1_{\text{ROUGE-L}} = 2 \times \frac{P_{LCS} \times R_{LCS}}{P_{LCS} + R_{LCS}} $$

\subsubsection*{BLEU (Bilingual Evaluation Understudy)}
BLEU is a metric for evaluating the quality of text which has been machine-translated from one natural language to another. Quality is considered to be the correspondence between a machine's output and that of a human. Scores are calculated for individual translated segments—generally sentences—by comparing them with a set of good quality reference translations.

The BLEU score is calculated as:
$$ \text{BLEU} = BP \cdot \exp\left(\sum_{n=1}^{N} w_n \log p_n\right) $$
where:
\begin{itemize}
    \item $p_n$ is the n-gram precision for n-grams of size $n$. It is calculated as the number of n-grams in the candidate translation that appear in any reference translation, divided by the total number of n-grams in the candidate translation.
    $$ p_n = \frac{\sum_{C \in \{\text{Candidates}\}} \sum_{\text{n-gram} \in C} \text{Count}_{\text{clip}}(\text{n-gram})}{\sum_{C' \in \{\text{Candidates}\}} \sum_{\text{n-gram}' \in C'} \text{Count}(\text{n-gram}')} $$
    $\text{Count}_{\text{clip}}(\text{n-gram})$ is the maximum number of times an n-gram occurs in any single reference translation, clipped to its count in the candidate translation.
    \item $N$ is the maximum n-gram length (typically 4).
    \item $w_n$ are positive weights for each $p_n$, typically uniform weights $w_n = 1/N$.
    \item $BP$ is the Brevity Penalty, which penalizes generated translations that are shorter than the closest reference length.
    $$ BP = \begin{cases} 1 & \text{if } c > r \\ e^{(1 - r/c)} & \text{if } c \le r \end{cases} $$
    where $c$ is the length of the candidate translation, and $r$ is the effective reference corpus length (usually the length of the reference sentence closest to the candidate sentence's length).
\end{itemize}

\subsubsection*{Edit Distance (Levenshtein Distance)}
Edit distance measures the similarity between two strings by counting the minimum number of single-character edits (insertions, deletions, or substitutions) required to change one string into the other. We use the Levenshtein distance.
Given two strings $a$ of length $m$ and $b$ of length $n$, the Levenshtein distance $lev_{a,b}(i,j)$ between the first $i$ characters of $a$ and the first $j$ characters of $b$ is given by:
$$ lev_{a,b}(i,j) = \begin{cases} j & \text{if } i=0 \\ i & \text{if } j=0 \\ lev_{a,b}(i-1, j-1) & \text{if } a_i = b_j \\ 1 + \min \begin{cases} lev_{a,b}(i-1, j) \\ lev_{a,b}(i, j-1) \\ lev_{a,b}(i-1, j-1) \end{cases} & \text{if } a_i \neq b_j \end{cases} $$
The edit distance between the full strings $a$ and $b$ is $lev_{a,b}(m,n)$. A lower edit distance indicates higher similarity. For normalization, it can be divided by the length of the longer string:
$$ \text{Normalized Edit Distance} = \frac{lev_{a,b}(m,n)}{\max(m,n)} $$
Similarity can then be expressed as $1 - \text{Normalized Edit Distance}$.

\subsubsection*{BERT Score (F1)}
BERT Score leverages the pre-trained contextual embeddings from BERT to compare the similarity between a candidate sentence and a reference sentence. It computes cosine similarity between the BERT embeddings of tokens in the candidate and reference sentences.

For each token $x_i$ in the reference sentence $X = \langle x_1, ..., x_k \rangle$ and each token $y_j$ in the candidate sentence $Y = \langle y_1, ..., y_m \rangle$, contextual embeddings $\mathbf{x}_i$ and $\mathbf{y}_j$ are obtained.

Recall ($R_{\text{BERT}}$) is calculated by matching each token in the reference to the most similar token in the candidate:
$$ R_{\text{BERT}} = \frac{1}{k} \sum_{i=1}^{k} \max_{j=1}^{m} \mathbf{x}_i^T \mathbf{y}_j $$
Precision ($P_{\text{BERT}}$) is calculated by matching each token in the candidate to the most similar token in the reference:
$$ P_{\text{BERT}} = \frac{1}{m} \sum_{j=1}^{m} \max_{i=1}^{k} \mathbf{x}_i^T \mathbf{y}_j $$
The BERT Score F1 is then the harmonic mean of $P_{\text{BERT}}$ and $R_{\text{BERT}}$:
$$ F1_{\text{BERT}} = 2 \frac{P_{\text{BERT}} \cdot R_{\text{BERT}}}{P_{\text{BERT}} + R_{\text{BERT}}} $$
Importance weighting for tokens can also be applied using inverse document frequency (IDF) scores, but the above formulas represent the basic BERT Score.

\begin{figure}[H]
    \centering
    \begin{subfigure}[b]{0.48\linewidth}
        \centering
        \includegraphics[width=\linewidth]{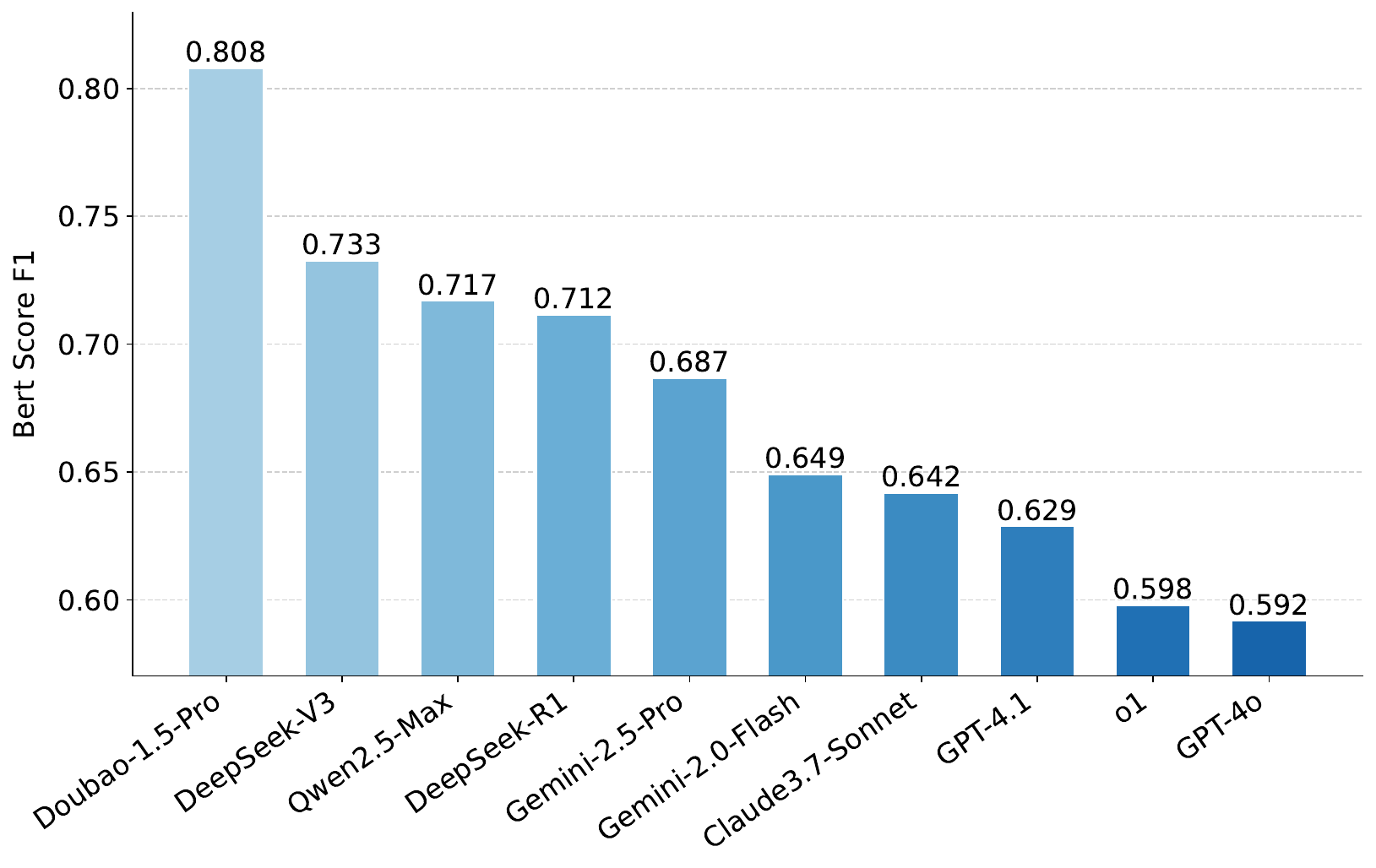}
        \caption{BERTScore for Content Retrieval.}
        \label{fig:enter-label-1}
    \end{subfigure}
    \hfill 
    \begin{subfigure}[b]{0.48\linewidth}
        \centering
        \includegraphics[width=\linewidth]{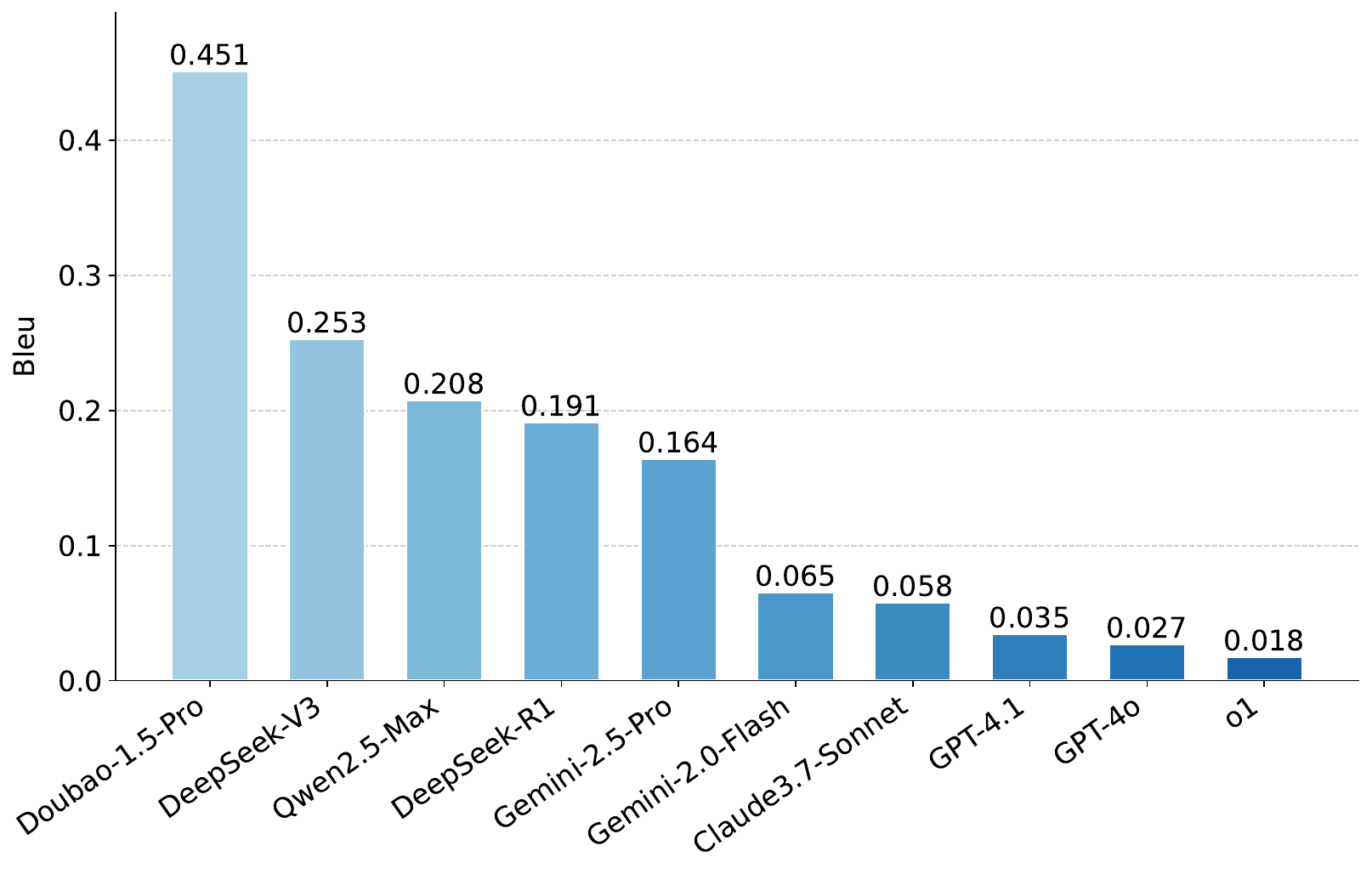}
        \caption{BLEU for Content Retrieval.}
        \label{fig:enter-label-2}
    \end{subfigure}
    
    \vspace{1em}

    \begin{subfigure}[b]{0.48\linewidth}
        \centering
        \includegraphics[width=\linewidth]{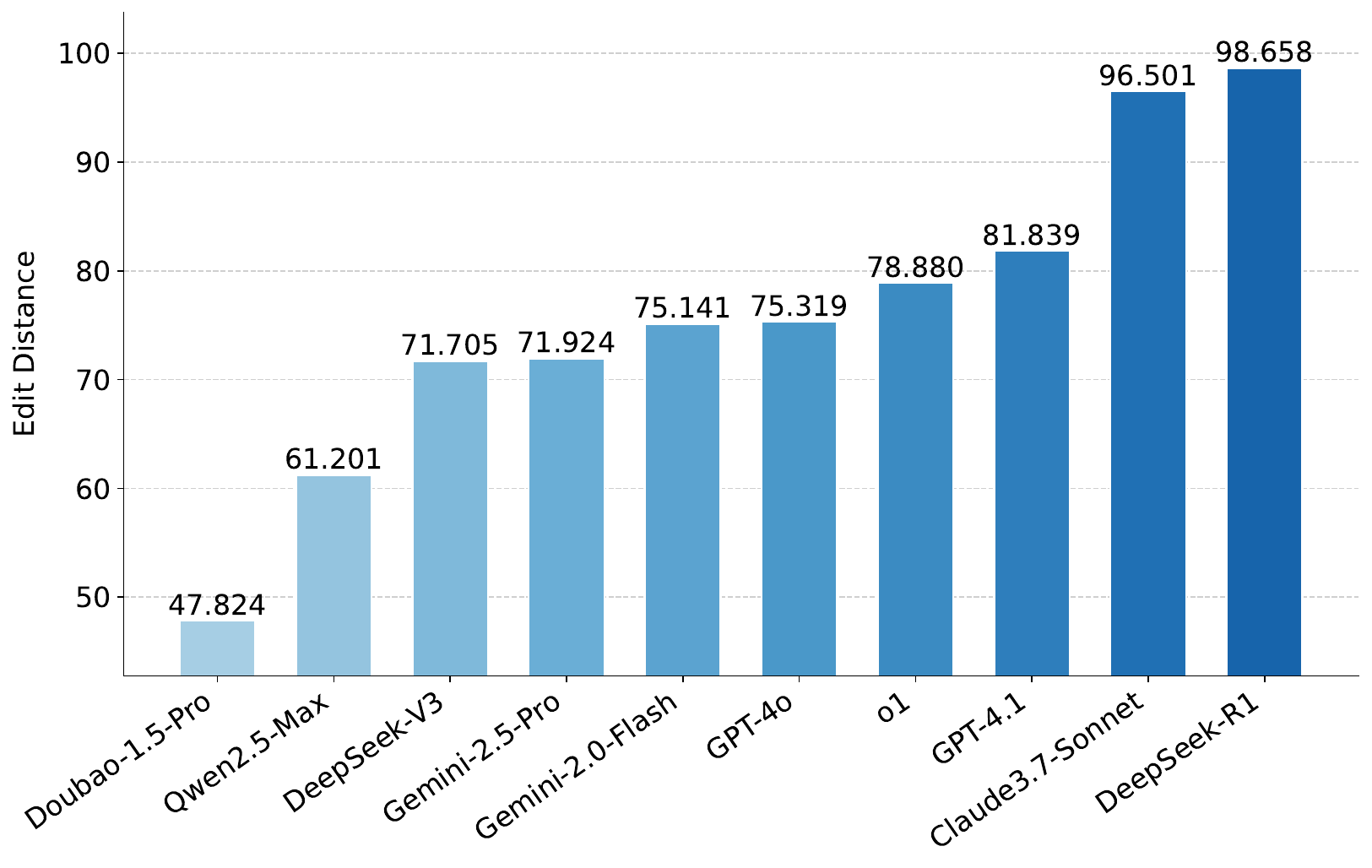}
        \caption{Edit distance for Content Retrieval.}
        \label{fig:enter-label-3}
    \end{subfigure}
    \hfill 
    \begin{subfigure}[b]{0.48\linewidth}
        \centering
        \includegraphics[width=\linewidth]{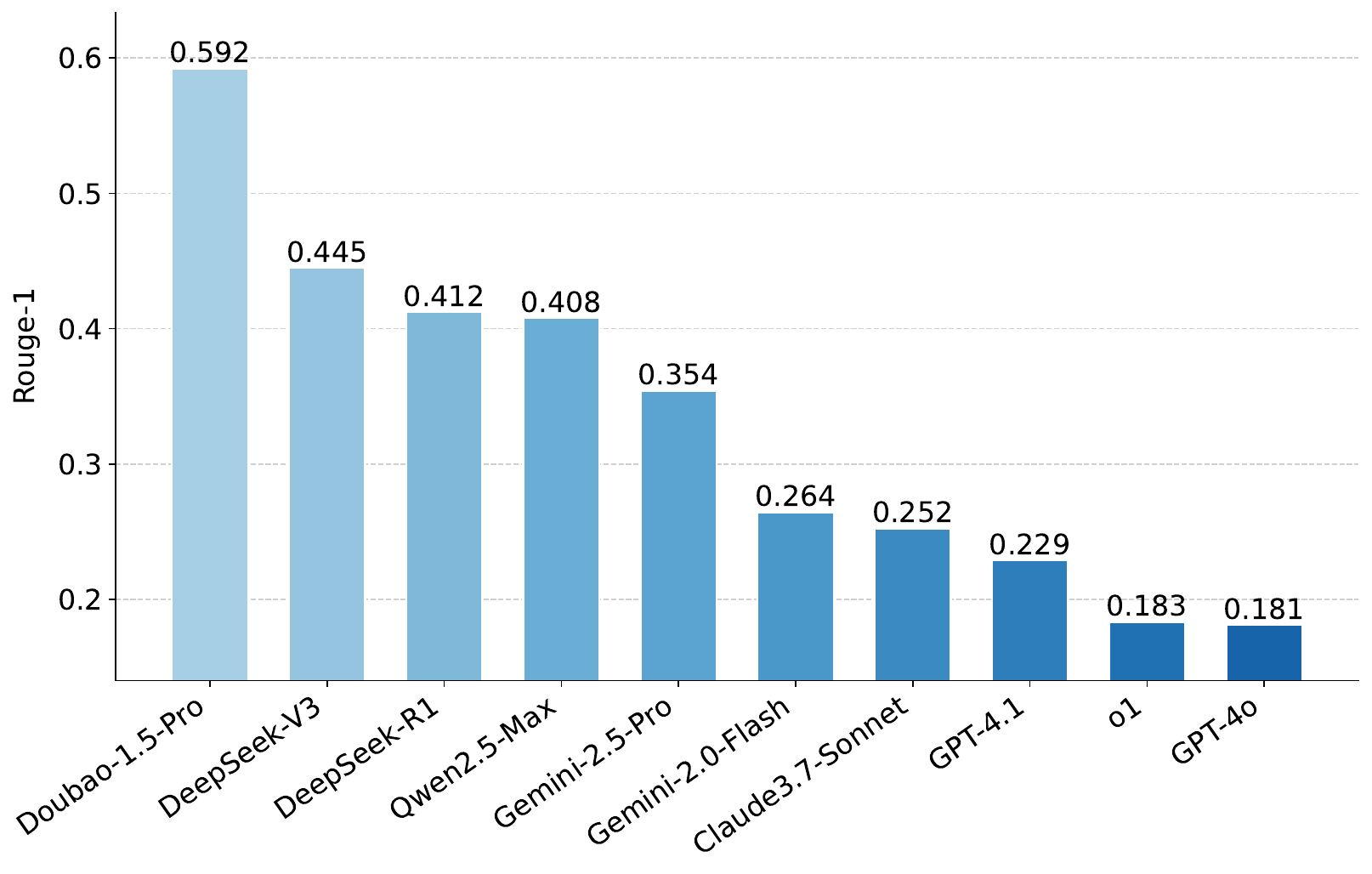}
        \caption{ROUGE-1 for Content Retrieval.}
        \label{fig:enter-label-4}
    \end{subfigure}

    \vspace{2em}
    
    \begin{subfigure}[b]{0.48\linewidth}
        \centering
        \includegraphics[width=\linewidth]{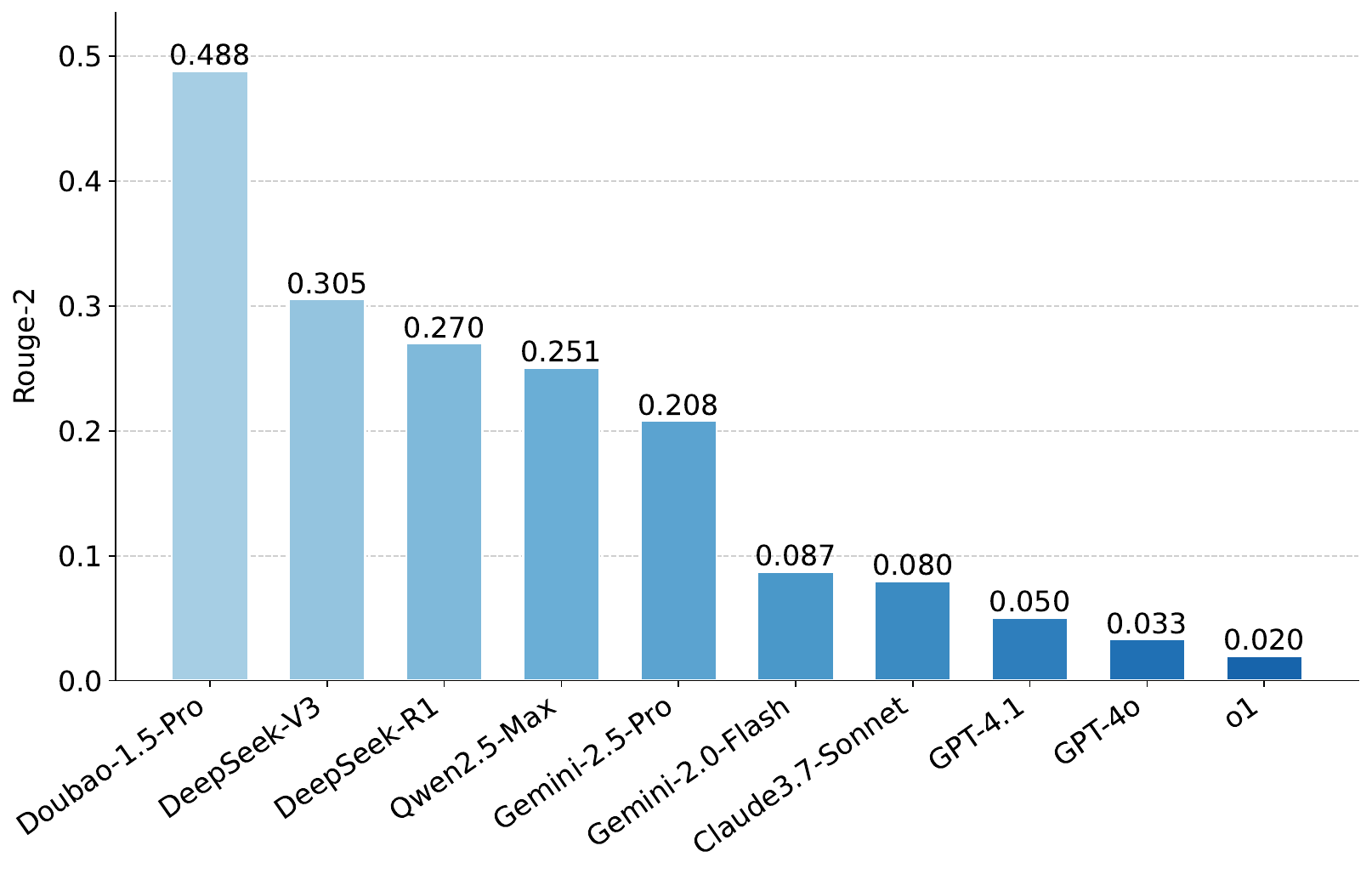}
        \caption{ROUGE-2 for Content Retrieval.}
        \label{fig:enter-label-5}
    \end{subfigure}
    \hfill 
    \begin{subfigure}[b]{0.48\linewidth}
        \centering
        \includegraphics[width=\linewidth]{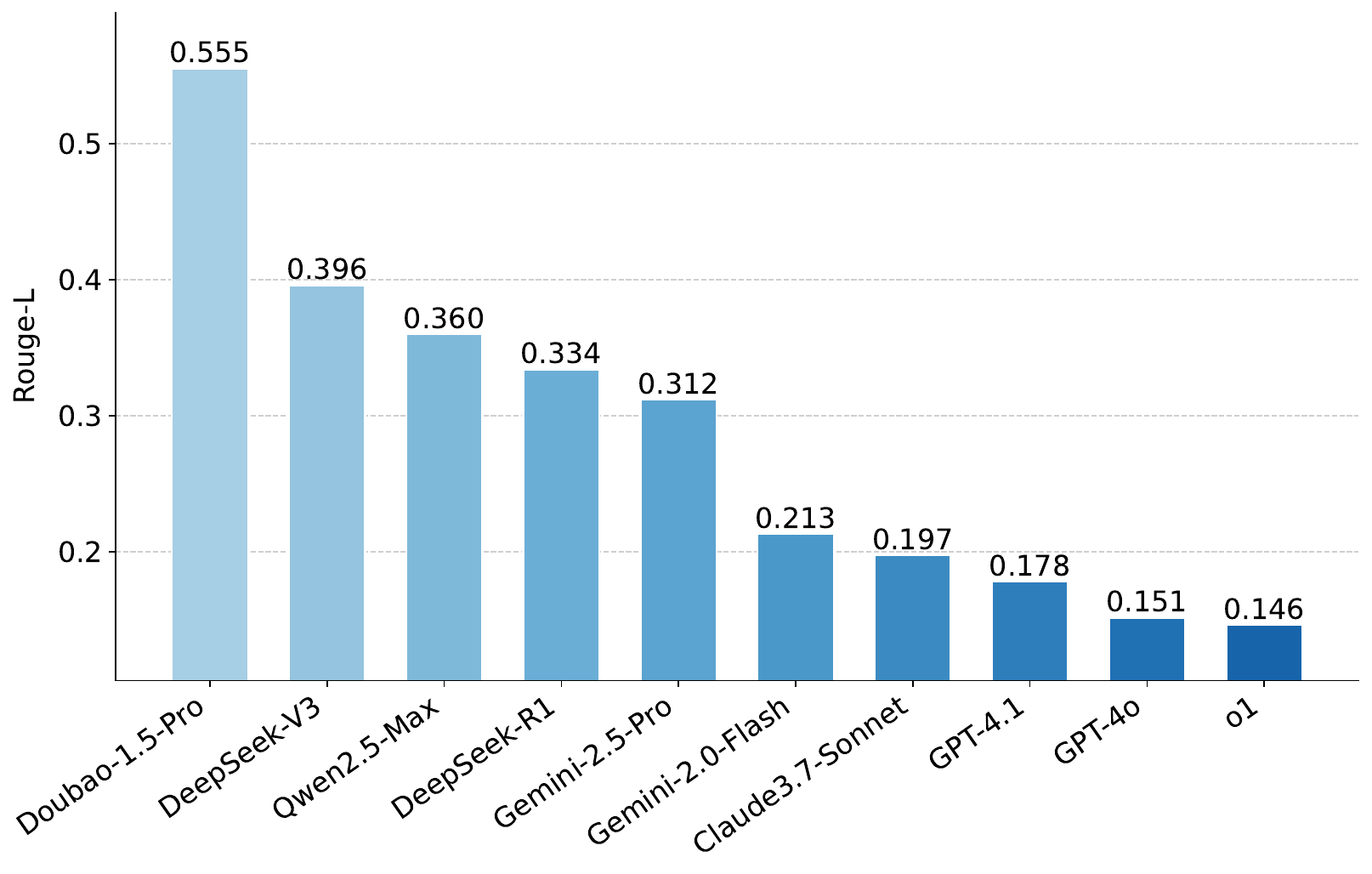}
        \caption{ROUGE-L for Content Retrieval.}
        \label{fig:enter-label-6}
    \end{subfigure}
    \caption{Overall Content Retrieval Performance Metrics. These figures show the performance of ten LLMs on various metrics for the content retrieval task. Higher scores are better for BERTScore, BLEU, ROUGE-1/2/L, while lower is better for Edit Distance.}
    \label{fig:content}
\end{figure}

The six sub-plots in \Cref{fig:content} report BERTScore, BLEU, Edit-distance, ROUGE-1/2/L on the same ten LLMs.
Three high-level findings emerge.

\textbf{(1) A single model—\emph{Doubao-1.5-Pro}—dominates every metric.}
Doubao achieves the highest semantic similarity (BERTScore 0.808) and the lowest character-level error (Edit-distance 47.8). Its margin is even more striking on \emph{lexical} measures: BLEU is 0.451, almost $2\times$ the second-best model (DeepSeek-V3, 0.253), and ROUGE-2/L are likewise $\ge 10$~pp ahead. Such uniform leadership suggests that Doubao has memorised large portions of the statutes verbatim and can reproduce them with minimal distortion.

\textbf{(2) Chinese–origin models again occupy the top tier, but the ranking reshuffles.}
DeepSeek-V3, Qwen-2.5-Max and DeepSeek-R1 occupy places 2–4 on all metrics, mirroring the ID-Retrieval trend that Chinese oriented LLMs profit from denser exposure to Chinese legal corpora. However, unlike ID-Retrieval—where DeepSeek-R1 overtook Doubao on fine-grained indices—here DeepSeek-R1 is only fourth. This inversion implies a trade-off: DeepSeek-R1 has stronger \emph{structural localisation} skills, whereas Doubao recalls the surface wording better.

\textbf{(3) Large, general-purpose Western models lag sharply.}
GPT-4.1, GPT-4o and o1 occupy the bottom three slots on every metric except Edit-distance, where Claude-3.7-Sonnet is marginally worse. On lexical metrics the gap is dramatic: GPT-4o’s BLEU (0.018) and ROUGE-2 (0.033) are an order of magnitude below Doubao. Semantic similarity narrows the gap (BERTScore 0.592 vs 0.808), indicating that these models can paraphrase the gist but rarely reproduce the exact statutory language—a behaviour consistent with broader observations that GPT-4 prioritises paraphrase over verbatim recall.

\medskip
\noindent\textbf{Lexical vs.\ semantic metrics.}
All models score higher on BERTScore than on BLEU/ROUGE, confirming heavy paraphrasing. The disparity is smallest for Doubao (0.808 vs 0.451) and grows for lower-ranked models (GPT-4o: 0.592 vs 0.018). Hence Doubao’s advantage is partly raw memorisation, while the others rely on semantic re-generation that incurs lexical penalties and legal-risking rewrites.

\medskip
\noindent\textbf{Absolute difficulty remains high.}
Even the best Edit-distance of 47.8 implies roughly one character error per eleven characters, and the top ROUGE-L of 0.555 means that almost half of the longest common subsequence is still missing or altered. Thus, like ID-Retrieval, Content Retrieval is \emph{far from solved}; precise, court-grade quotations cannot yet be trusted to any public LLM.

\medskip
\noindent\textbf{Cross-task correlation.}
Spearman correlation between article-level ID accuracy and BERTScore is 0.83, but drops to 0.42 at subparagraph level—models that find the right clause do not necessarily quote it correctly. The dissociation is clearest for DeepSeek-R1 (excellent locator, moderate quoter) versus Doubao (excellent quoter, slightly weaker locator), suggesting complementary strengths that could be combined in a pipeline system.

These findings underscore the need for retrieval-augmented generation or explicit citation verification before deploying LLMs in professional legal workflows.

\subsection{Ablation of LLM Judge}\label{app.llm_judge}

\begin{table*}[h]
\caption{Legal Case Reasoning performance as evaluated by an LLM judge (DeepSeek-R1). The maximum score for Overall Rating is 100, while the five individual aspects (Reasoning, Knowledge, Structure, Clarity, Conciseness) are scored out of 20. Note that the overall rating is not a simple sum of these aspects. For the aspects, qualitative ratings of \emph{Good}, \emph{Normal}, \emph{Bad} were converted to scores of 20.0, 10.0, and 0.0, respectively. All presented results are an average across 254 cases.}
\label{tab:reasoning_performance_deepseek_r1_judge} 
\centering
\resizebox{\textwidth}{!}{%
\begin{tabular}{lcccccc}
\toprule
\textbf{Model Name} & \textbf{Overall Rating} & \textbf{Reasoning} & \textbf{Knowledge} & \textbf{Structure} & \textbf{Clarity} & \textbf{Conciseness} \\
\midrule
o1 & 65.78 & 10.00 & 0.00 & 10.00 & 10.00 & 10.00 \\
GPT-4o & 71.00 & 5.00 & 5.00 & 10.00 & 5.00 & 15.00 \\
Gemini-2.0-Flash & 74.01 & 12.69 & 10.00 & 19.23 & 18.08 & 17.31 \\
Doubao-1.5-Pro & 75.39 & 5.00 & 5.00 & 17.50 & 15.00 & 17.50 \\
Claude-3.7-Sonnet & 77.33 & 6.67 & 6.67 & 13.33 & 10.00 & 16.67 \\
Qwen-2.5-Max & 79.24 & 8.57 & 5.71 & 17.14 & 14.29 & 14.29 \\
GPT-4.1 & 80.16 & 14.29 & 11.43 & 18.57 & 20.00 & 10.00 \\
DeepSeek-V3 & 83.37 & 15.00 & 7.50 & 17.50 & 17.50 & 20.00 \\
DeepSeek-R1 & 83.56 & 18.33 & 11.67 & 20.00 & 20.00 & 16.67 \\
Gemini-2.5-Pro & 84.35 & 18.67 & 14.67 & 19.33 & 18.67 & 15.33 \\
\bottomrule
\end{tabular}%
}
\end{table*}

\end{CJK} 
\end{document}